\documentclass[11pt,a4paper]{article}
\usepackage[T1]{fontenc}
\usepackage[utf8]{inputenc}
\usepackage{lmodern,microtype,amsmath,amssymb,amsthm,mathtools,bm,mathrsfs}
\usepackage[margin=26mm]{geometry}
\usepackage{graphicx,booktabs,array,longtable,enumitem,float,placeins}
\usepackage[colorlinks=true,allcolors=blue!45!black]{hyperref}
\usepackage{xcolor,cleveref,fancyhdr}
\usepackage{tikz,tikz-3dplot,pgfplots}
\usetikzlibrary{arrows.meta,calc,decorations.pathreplacing,decorations.markings,patterns,positioning,backgrounds}
\pgfplotsset{compat=1.18}
\definecolor{deep}{RGB}{25,64,92}
\definecolor{teal}{RGB}{9,119,127}
\definecolor{coral}{RGB}{191,78,62}
\definecolor{golden}{RGB}{177,130,46}
\definecolor{soft}{RGB}{230,244,248}
\hypersetup{pdftitle={Nonreciprocal Spherical Attention and Retrieval: Sharp Coarsening Clocks and Certified Score Correction},pdfauthor={Faruk Alpay and Taylan Alpay},pdfsubject={Compressed dense retrieval and metastable spherical attention},pdfkeywords={nonreciprocal attention, dense retrieval, product quantization, skew-symmetric calibration, spherical clustering, metastability, formal verification}}
\newtheorem{theorem}{Theorem}[section]
\newtheorem{lemma}[theorem]{Lemma}
\newtheorem{proposition}[theorem]{Proposition}
\newtheorem{corollary}[theorem]{Corollary}
\theoremstyle{definition}
\newtheorem{definition}[theorem]{Definition}
\newtheorem{problem}[theorem]{Open problem}
\theoremstyle{remark}\newtheorem{remark}[theorem]{Remark}
\newcommand{\R}{\mathbb R}\newcommand{\sphere}{\mathbb S}
\newcommand{\eps}{\varepsilon}\newcommand{\dd}{\mathrm d}
\newcommand{\T}{\mathsf T}
\newcommand{\cP}{\mathcal P}\newcommand{\cC}{\mathcal C}\newcommand{\cM}{\mathcal M}

\newcommand{\ang}{\operatorname{dist}_{\sphere}}

\newcommand{\SciBytes}{430,016}
\newcommand{\NFBytes}{330,816}

\title{\textbf{Nonreciprocal Spherical Attention and Retrieval:}\\
Sharp Coarsening Clocks and Certified Score Correction}
\author{\begin{tabular}{c@{\hspace{1.2em}}c}
Faruk Alpay\thanks{Correspondence: \href{mailto:alpay@lightcap.ai}{alpay@lightcap.ai}.} & Taylan Alpay\\[3pt]
\small Department of Computer Engineering & \small Department of Aerospace\\
\small Bahçeşehir University & \small University of Turkish Aeronautical Association\\
\small Istanbul, Türkiye & \small Ankara, Türkiye\\
\small\href{mailto:faruk.alpay@bahcesehir.edu.tr}{faruk.alpay@bahcesehir.edu.tr} & \small\href{mailto:s220112602@stu.thk.edu.tr}{s220112602@stu.thk.edu.tr}
\end{tabular}}
\date{}
\begin{document}
\maketitle
\begin{abstract}
We study nonreciprocal bilinear interactions as a common structure for
spherical coarsening and compressed nearest-neighbor retrieval. For normalized
attention, we establish a sharp two-cluster metastability exponent and extend
it to growing particle systems under a subexponential block-mass ratio condition,
without a particle-count restriction.
A one-sided entrance argument determines the second-event exponent in an
explicit three-body flow. For finite compression, mass centering cancels the
first-order error and yields a sharp second-order attention-output bound,
including the correction required by spherical normalization. We formulate
skew score calibration as a strictly convex moment problem, derive its unique
matrix normal equation, and give residual-based optimization error bounds.
A canonical low-rank projection provides bounded, well-conditioned query
updates with an explicit index-memory cost. Experiments combine semantic
retrieval ablations with one- and ten-million-vector visual descriptor indices,
recomputed cosine ground truth, and native FAISS PQ/OPQ search. On SIFT1M,
three-seed native-L2 hit@10 rises from 0.8926 to 0.8948 with norm-aware skew
margin calibration, while symmetric calibration reaches 0.8996. At ten
million vectors, IVF-PQ reaches 0.8827 hit@10 at 0.1794 ms per query in a
four-thread batch benchmark; validation selects no skew update. A separate
ten-million-key attention-memory experiment evaluates centered aggregation
against exact streamed output on 1,024 held-out queries. Matrix-basis solves,
automatic differentiation, finite differences, interval arithmetic, and
full-flow checks complement nine kernel-checked algebra lemmas. The results
support precise clock and approximation guarantees, without establishing
universal retrieval superiority from directionality.
\end{abstract}
\section{Introduction}
Dense retrieval ranks document embeddings $x_i\in\sphere^{d-1}$ by
$q^\top x_i$. Compression replaces those vectors by a smaller index; success
requires ranking quality at a stated byte budget, not merely visually
separated clusters. This distinction is essential because attention-induced
clustering is already well studied~\cite{GLPR,Perspective,GKPR,Rigollet}, and
representation collapse can harm useful information~\cite{Dong}.

We investigate three questions: whether directionality helps retrieval at a
fixed budget, whether it attenuates a specified representation perturbation,
and what mathematical guarantees survive nonreciprocity. The initial
centroid-only hypothesis fails against competitive quantization. That
failure motivates a second, explicit hypothesis: constrained skew score
calibration may correct task-relevant quantization errors without the
ill-conditioning allowed by a general symmetric metric.

The analysis provides three connected results: a sharp metastability
exponent controlled by block mass even under exponential particle growth;
a second-order attention compression estimate obtained by mass centering;
and a unique skew score-correction equation with a computable optimization
residual bound. Experiments evaluate both semantic retrieval and geometric
nearest-neighbor search, including a ten-million-vector visual-descriptor index.
The scalar probability inequalities and standard convex optimization tools
used in the derivations are attributed to their established sources.

\section{Related Work}
\textbf{Spherical attention and clustering.}
The cluster formation, cone-collapse, and metastability results
of~\cite{GLPR,Perspective,GKPR} are the closest mathematical predecessors.
The cone argument in~\cite[Lemma 4.2 and its proof]{Perspective} explicitly
extends beyond symmetric query--key matrices. The acute consensus estimate used here is a quantitative specialization
of that established positive-interaction argument. Directional mean
shift~\cite{DMS,DMSEM} and diffusion condensation~\cite{Diffusion} supply
related density-based and time-inhomogeneous clustering formulations.
Directed-sphere consensus has its own established
literature~\cite{Markdahl,Lifting}.

\textbf{Directionality and skew geometry.}
Symmetry and directionality during transformer training are analyzed
in~\cite{Saponati}. Routing/filtering separation~\cite{Routing} and
orthogonal attention~\cite{OSA} use skew structure for different objectives.
The latter exponentiates a token interaction operator; our finite retrieval
adapter perturbs query--document scores. These token-space operators differ from the descriptor-space correction
optimized here against compressed retrieval scores. Symmetric spectral selection~\cite{KY}
has different assumptions from the directed barrier clocks studied here.

\textbf{Compression and asymmetric retrieval.}
Two-level softmax size and dispersion corrections are studied
in~\cite{TwoLevel}. Here the aggregation analysis controls a vector-valued
attention output with skew bilinear scores and explicitly accounts for
spherical normalization, rather than claiming size correction as new.
Product quantization~\cite{PQ}, anisotropic vector
quantization~\cite{AVQ}, query-aware quantization~\cite{QAQ}, and
reduced-rank score regression~\cite{LoRANN} are essential comparators.
Joint query-encoder and quantizer optimization is studied in~\cite{JPQ}.
Our correction leaves the descriptor extractor fixed, constrains the score
operator to a skew subspace, and solves a convex moment equation before
rank compression. Low-rank score approximation and query-aware compression
are established components. Asymmetric hashing~\cite{ALSH,Neyshabur} also shows
that asymmetry requires a precise task setting; its presence alone does not
establish an advantage. Learned rotations for quantization predate this
work~\cite{ITQ}. Token merging~\cite{ToMe} supplies a practical precedent for
mass-aware aggregation, but reducing tokens inside a transformer and
compressing an external retrieval index are distinct tasks.

The structural Gram identities and exact duplicate aggregation are supporting
lemmas. The analytic contribution concerns shielded directed event selection
and a two-cluster metastability exponent under a vanishing-skew restriction.
For compressed retrieval, the contribution is a constrained score-correction
formulation with explicit conditioning, optimization residual, and attention
aggregation error bounds. These results distinguish the scope of the present
model from general spherical consensus, generic asymmetric retrieval, and
unconstrained low-rank regression.

\section{Spherical Interaction Model and Retrieval Formulation}
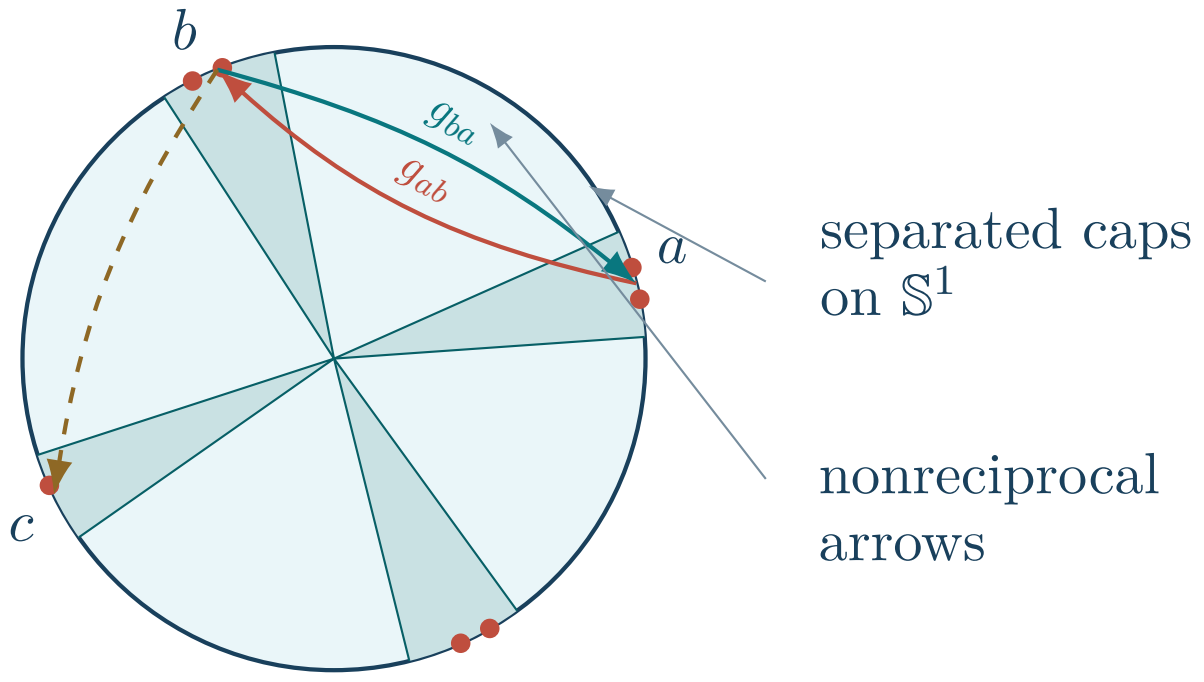
\begin{figure}[H]
\centering
\begin{tikzpicture}[scale=1.02,>=Latex]
\def\rad{2.02}
\fill[soft!85] (0,0) circle (\rad);
\draw[deep,thick] (0,0) circle (\rad);
\foreach \c/\lo/\hi in {14/4/24,112/101/123,206/198/215,294/284/306}{
  \fill[teal!22] (0,0) -- (\lo:\rad) arc[start angle=\lo,end angle=\hi,radius=\rad] -- cycle;
  \draw[teal!80!black,thin] (0,0)--(\lo:\rad) (0,0)--(\hi:\rad);
}
\foreach \ang/\col in {11/coral,17/coral,111/coral,117/coral,204/coral,300/coral,294/coral}{
  \fill[\col] (\ang:\rad) circle (1.8pt);
}
\coordinate (a) at (14:\rad);\coordinate (b) at (112:\rad);\coordinate (c) at (206:\rad);
\draw[coral,thick,->] (a) to[bend left=15] node[above,sloped,scale=.77]{$g_{ab}$} (b);
\draw[teal,thick,->] (b) to[bend left=12] node[above,sloped,scale=.77]{$g_{ba}$} (a);
\draw[golden!80!black,thick,dashed,->] (b) to[bend right=12] (c);
\node[above right,deep] at (a) {$a$};\node[above left,deep] at (b) {$b$};\node[below left,deep] at (c) {$c$};
\node[deep,align=left] at (4.36,.63) {separated caps\\[-1pt]on $\sphere^1$};
\draw[deep!60,->] (2.8,.5) -- (1.65,1.12);
\node[deep,align=left] at (4.25,-.96) {nonreciprocal\\[-1pt]arrows};
\draw[deep!60,->] (2.8,-.78)--(1.01,1.53);
\end{tikzpicture}
\caption{Our geometric notation: caps on the unit circle, with directed interactions and unequal barriers. The cap construction is visually motivated by Figure~1 of~\cite{GKPR}; this drawing is original.}
\label{fig:caps}
\end{figure}

For positive masses $m_j$, skew $A^\top=-A$, $M=I+\eps A$, and $\beta>0$,
the continuous model is
\begin{equation}\label{eq:model-main}
 \dot u_i=(I-u_iu_i^\top)\sum_jp_{ij}u_j,
 \qquad p_{ij}=\frac{m_j e^{\beta u_i^\top M u_j}}
                         {\sum_km_k e^{\beta u_i^\top M u_k}}.
\end{equation}
For $A=0$ and equal masses, \eqref{eq:model-main} is the normalized
spherical self-attention flow studied in~\cite{GLPR,GKPR}; general
query--key bilinear scores are formulated in~\cite[Section~2.2]{Perspective}.
The exponential dot-product kernel is also related to directional mean
shift~\cite{DMS,DMSEM}, although the projected continuous dynamics here
are distinct from its discrete mode-seeking iteration.
Its directed barrier is $g_{ij}=1-u_i^\top M u_j$, with
$g_{ij}-g_{ji}=-2\eps u_i^\top Au_j$. Appendix~\ref{sec:gram-geometry} gives
closed ordinary/skew Gram dynamics, a shielded first-event theorem, an
explicit inversion of nearest-pair order, and rotating triangles that
exclude unconditional coarsening. Self-edges are explicitly excluded from
the corridor's positive external-barrier condition.
Figure~\ref{fig:caps} fixes the geometric notation and distinguishes the
two directed barriers between neighboring caps.

The same skew form measures cycle directionality. For
$H_{ij}=u_i^\top Au_j$, the three-cycle log affinity is
$2\beta\eps(H_{ab}+H_{bc}+H_{ca})$.
For the planar rotation $A=J$, it equals $-4\beta\eps$ times the signed
triangle area, as shown in Figure~\ref{fig:area}.
Masses and row normalizers cancel exactly; the full derivation is in
Appendix~\ref{sec:gram-geometry}.
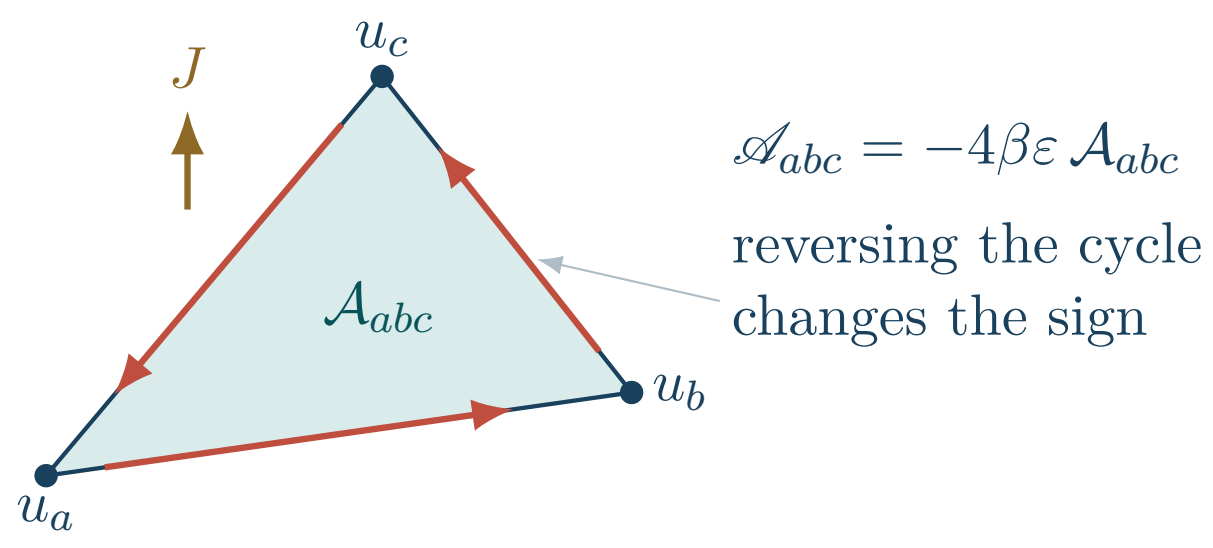
\begin{figure}[H]
\centering
\begin{tikzpicture}[scale=1.1,>=Latex]
\coordinate (a) at (-1.8,-.7);\coordinate (b) at (1.72,-.2);\coordinate (c) at (.22,1.7);
\fill[teal!15] (a)--(b)--(c)--cycle;
\draw[deep,thick] (a)--(b)--(c)--cycle;
\draw[-{Latex[length=2.8mm]},coral,very thick] ($(a)!.10!(b)$)--($(a)!.80!(b)$);
\draw[-{Latex[length=2.8mm]},coral,very thick] ($(b)!.13!(c)$)--($(b)!.78!(c)$);
\draw[-{Latex[length=2.8mm]},coral,very thick] ($(c)!.12!(a)$)--($(c)!.80!(a)$);
\foreach \p/\loc/\lab in {a/below/$u_a$,b/right/$u_b$,c/above/$u_c$}{\fill[deep] (\p) circle (2pt);\node[\loc,deep] at (\p) {\lab};}
\draw[golden!80!black,->,very thick] (-.95,0.9)--(-.95,1.5) node[above] {$J$};
\node[teal!70!black] at (.20,.31) {$\mathcal A_{abc}$};
\node[align=left,deep,anchor=west] at (2.2,.75) {$\mathscr A_{abc}=-4\beta\eps\,\mathcal A_{abc}$\\[5pt] reversing the cycle\\ changes the sign};
\draw[deep!35,->] (2.25,.35)--(1.15,.60);
\end{tikzpicture}
\caption{The skew interaction measures the directed signed area of a triangle. The figure represents a two-dimensional linear plane; the vertices may also be constrained to a unit circle.}
\label{fig:area}
\end{figure}

The first ML pilot uses a finite algorithm:
$u_i^+=\operatorname{normalize}(u_i+h(I-u_iu_i^\top)\sum_jp_{ij}u_j)$.
We take $\beta=12$, $h=0.25$, and two or four steps selected on validation.
This is not an exact ODE solver and does not invoke a recursive-merger
theorem. Spherical Lloyd clustering of the final vectors selects a partition
with $K\in\{128,256,512\}$; prototypes average the \emph{original} vectors.
Document scores are prototype scores expanded through an assignment array.
All documents in a cluster therefore tie, resolved by a fixed lexicographic
ID order. The pilot also compares a positive-pair-moment skew learned only
from fitting queries, its sign reversal, random skew, reciprocal flow, and
ordinary spherical clustering. A supervised symmetric-moment control was
added after the initial pilot and is labeled exploratory.

\begin{proposition}[A centroid-only ranking obstruction]\label{prop:tie}
If distinct unit vectors $x_i,x_j$ share a reconstructed vector, there is a
unit query for which their exact score gap is $\|x_i-x_j\|$ while their
compressed scores tie. No centroid-only representation of that cluster
preserves every query's strict pairwise ordering.
\end{proposition}
\begin{proof}
Take $q=(x_i-x_j)/\|x_i-x_j\|$. The exact score difference is the stated
norm; the reconstructed difference is zero. A fixed tie-break cannot recover
both this ordering and its reversal under $-q$.
\end{proof}
This is a standard distortion obstruction applied to the pilot. Improving
the dynamics alone cannot restore discarded within-cluster information.

\section{Certified Nonreciprocal Score Correction}\label{sec:certified}
Let $X,\widehat X\in\mathbb R^{N\times d}$ contain original and compressed
document vectors, and let $Q\in\mathbb R^{s\times d}$ contain fitting queries.
Score approximation as multivariate regression is established in
LoRANN~\cite[Sections~3.1--3.2]{LoRANN}; score-sensitive quantizer losses
are developed in~\cite[Section~3, Theorem~3.2]{AVQ} and~\cite{QAQ}.
Here the codebook is fixed and the regression variable is restricted to a
skew query-side residual. The resulting retrieval distortion is
\begin{equation}\label{eq:ridge-score}
 \mathcal J(A)=\frac{\|Q A\widehat X^\top-Q(X-\widehat X)^\top\|_F^2}{2sN}
                 +\frac\lambda2\|A\|_F^2,\qquad A^\top=-A,
 \quad\lambda>0.
\end{equation}
Thus the target is the uncompressed score, rather than a geometric
reconstruction loss that weights every direction equally. Define
\[
 C=Q^\top Q/s,\quad H=\widehat X^\top\widehat X/N,\quad
 D=(X-\widehat X)^\top\widehat X/N,\quad
 E=(X-\widehat X)^\top(X-\widehat X)/N,
\]
and write $\Pi_-(B)=(B-B^\top)/2$. These moments are sufficient to fit the
correction without storing a query-by-document score matrix.

\begin{theorem}[Unique calibrated skew solution and residual certificate]\label{thm:certified-skew}
Problem~\eqref{eq:ridge-score} has a unique minimizer $A_*$, characterized by
\begin{equation}\label{eq:skew-normal}
 \mathcal T(A_*):=\Pi_-(C A_* H)+\lambda A_*=\Pi_-(CD).
\end{equation}
On the real skew subspace, $\mathcal T$ is self-adjoint and positive definite,
with eigenvalues in $[\lambda,L]$, where
$L=\lambda+\|C\|_{\rm op}\|H\|_{\rm op}$. For any skew approximation $B$,
put $R_B=\Pi_-(CBH-CD)+\lambda B$. Then
\begin{align}
 \|B-A_*\|_F&\le\|R_B\|_F/\lambda,\label{eq:cert-distance}\\
 0\le\mathcal J(B)-\mathcal J(A_*)&\le\|R_B\|_F^2/(2\lambda).
 \label{eq:cert-gap}
\end{align}
In particular $A_*=0$ if and only if $\Pi_-(CD)=0$. The bound certifies the
convex problem, not global optimality of a subsequent rank restriction.
\end{theorem}
\begin{proof}
Expanding the squared norm gives
\[
 2\mathcal J(A)=\operatorname{tr}(CAHA^\top)
       -2\langle CD,A\rangle_F+\operatorname{tr}(CE)+\lambda\|A\|_F^2.
\]
For skew variations $Z$, differentiation yields
$\dd\mathcal J(A)[Z]=\langle CAH-CD+\lambda A,Z\rangle_F$.
The orthogonal projection onto admissible variations gives
\eqref{eq:skew-normal}. For skew $Z,W$, symmetry of $C,H$ and cyclicity of
trace imply
$\langle Z,\mathcal T W\rangle_F=\langle\mathcal T Z,W\rangle_F$.
Moreover
\[
 \langle Z,\mathcal T Z\rangle_F
 =\|C^{1/2}ZH^{1/2}\|_F^2+\lambda\|Z\|_F^2.
\]
The lower bound is $\lambda\|Z\|_F^2$ and the upper bound is
$L\|Z\|_F^2$. Thus $\mathcal J$ is coercive and strictly convex on this
finite-dimensional subspace and has the unique stationary minimizer stated.
Set $F=B-A_*$. Then $R_B=\mathcal T F$, so
$\lambda\|F\|_F^2\le\langle F,R_B\rangle_F\le\|F\|_F\|R_B\|_F$,
which proves~\eqref{eq:cert-distance}, including $F=0$.
Exact quadratic expansion at the stationary point gives
$\mathcal J(B)-\mathcal J(A_*)=\langle F,\mathcal T F\rangle_F/2$.
Diagonalizing the self-adjoint operator on an orthonormal skew basis gives
$\langle F,\mathcal T F\rangle_F
=\langle R_B,\mathcal T^{-1}R_B\rangle_F\le\|R_B\|_F^2/\lambda$.
Finally, the normal equation at zero is equivalent to $\Pi_-(CD)=0$.
\end{proof}

The classical conjugate-gradient method~\cite{CG}
solves~\eqref{eq:skew-normal} using the matrix-vector
operation $A\mapsto\Pi_-(CAH)+\lambda A$. No inversion of a
$d^2$-by-$d^2$ matrix is required. General structured least squares and
Sylvester solvers precede this work~\cite{SkewLS,SylvesterOptimization}; the
result above specializes that machinery to quantized score distortion and
provides the residual used by the retrieval artifact. This formulation also
differs from unrestricted reduced-rank score regression~\cite{LoRANN} by
enforcing a Euclidean skew correction and its associated conditioning bound.

For a deployed rank-$2r$ correction, a real Schur decomposition groups the
nonzero skew blocks into
$\left(\begin{smallmatrix}0&a_j\\-a_j&0\end{smallmatrix}\right)$.
We retain the $r$ largest $|a_j|$ and clip each to a prescribed bound $b$.
The best-rank principle is the classical Eckart--Young result~\cite{EckartYoung},
applied to paired skew singular values.
This is the Frobenius-nearest skew matrix satisfying rank at most $2r$ and
spectral norm at most $b$: the singular values occur in equal pairs, the
unitarily invariant best-rank approximation retains the largest pairs, and
the scalar nearest point to $[-b,b]$ is clipping. If $\widetilde A$ is an
inexact full solution with residual at most $\eta$ and $B$ is this projection,
then
\[
 \mathcal J(B)-\mathcal J(A_*)
 \le\frac L2\left(\|B-\widetilde A\|_F+\frac\eta\lambda\right)^2.
\]
This follows from~\eqref{eq:cert-distance}, the triangle inequality, and the
quadratic upper bound. The full solver error and the rank-compression error
are therefore reported separately. Symmetric and unrestricted normal-equation
controls distinguish the skew constraint from generic regression capacity.

\subsection{Native-distance margin calibration}\label{sec:margin}
For unit original descriptors, cosine and squared-L2 exact rankings coincide.
After quantization, the reconstructed norms vary, so native L2 and inner-product
PQ rankings need not coincide. We therefore calibrate the score actually used
by the L2 index. For a fitting query $q_i$, let $x_i^+$ be its exact nearest
neighbor and let $x_{ik}^-$ be a compressed-search hard negative. Put
$\Delta_{ik}=\hat x_i^+-\hat x_{ik}^-$ and
\[
 t_{ik}=q_i^\top(x_i^+-x_{ik}^- -\Delta_{ik})
       +\tfrac12(\|\hat x_i^+\|^2-\|\hat x_{ik}^-\|^2).
\]
The target includes the reconstructed-norm term; dropping it calibrates a
different scoring rule. The finite margin objective is
\[
 \mathcal J_{\rm gap}(A)=\frac1{2sK}\sum_{i,k}
               (q_i^\top A\Delta_{ik}-t_{ik})^2
                        +\frac\lambda2\|A\|_F^2,\qquad A^\top=-A.
\]
Define $\mathcal L(A)_{ik}=q_i^\top A\Delta_{ik}/\sqrt{sK}$ and
$y_{ik}=t_{ik}/\sqrt{sK}$. Its unique minimizer solves
\[
 [\Pi_-\mathcal L^*\mathcal L+\lambda I]A_*=
                                      \Pi_-\mathcal L^*y.
\]
The proof of Theorem~\ref{thm:certified-skew} applies verbatim after replacing
the separable moment Hessian by $\mathcal L^*\mathcal L$: its quadratic form
is $\|\mathcal L Z\|_F^2+\lambda\|Z\|_F^2$, so existence, uniqueness, and
the residual error bound are unchanged. The implementation evaluates this
operator by grouped query--negative products without forming a design matrix.

To connect this surrogate to ranking, let $\gamma_{ik}=q_i^\top(x_i^+-x_{ik}^-)$
and let $\hat\gamma_{ik}$ be the calibrated native-L2 score gap. Expanding
$-\|q_i^\top(I+A)-\hat x\|^2/2$ and cancelling its query-only term gives
\[
 \hat\gamma_{ik}-\gamma_{ik}=q_i^\top A\Delta_{ik}-t_{ik}.
\]
For any $\gamma_0>0$, a deterministic finite-sample inequality is
\[
 \frac{\#\{(i,k):\hat\gamma_{ik}\le0\}}{sK}
 \le\frac{\#\{(i,k):\gamma_{ik}<\gamma_0\}}{sK}
        +\frac{2\mathcal J_{\rm gap}(A)}{\gamma_0^2}.
\]
Indeed, a reversed pair with $\gamma_{ik}\ge\gamma_0$ has squared gap error
at least $\gamma_0^2$; summing and retaining the nonnegative ridge term proves
the display. This bounds errors on the specified pairs and does not turn a
training-pair certificate into a test-distribution guarantee.

\section{Low-Rank Implementation and Index Budget}\label{sec:calibration}
Let $\hat x_i$ be a product-quantized reconstruction, with $m$ subspaces,
$L$ float32 codewords per subspace, and one uint8 code per document per
subspace, following the product-code representation of~\cite{PQ}.
The asymmetric-distance construction in~\cite[Section~III-A]{PQ} keeps
the query unquantized and evaluates it against compressed database vectors.
For squared-L2 search its ranking score, up to a query-only constant, is
\[
 s_{\rm L2}(q,i)=q^\top\hat x_i-\tfrac12\|\hat x_i\|^2.
\]
This standard identity explains why the norm term in
Section~\ref{sec:margin} cannot be dropped. Our correction changes the query
before the same compressed scorer; it does not replace PQ by a new codec.
The deployed scorer stores codebooks, codes, and two factors
$U,V\in\R^{d\times r}$:
\begin{equation}\label{eq:calibration}
 A=UV^\top-VU^\top,\qquad
 s_A(q,i)=q^\top(I+A)\hat x_i,\qquad \|A\|_{\rm op}\le\tfrac12.
\end{equation}
Here $\varepsilon$ is absorbed into $A$, and query and document spaces are
treated differently. The rank of $A$ is
at most $2r$, not $r$. In the semantic retrieval ablation, the factors are trained using the marginal
log-likelihood of judged relevant documents among all corpus documents,
with temperature 20 and a factor penalty $10^{-3}(\|U\|_F^2+\|V\|_F^2)$.
Only fitting queries supply supervision. Adam uses learning rate 0.01 for
40 epochs, batch size at most 64. The operator-norm projection is computed
exactly from a $2r$-dimensional factor core after each update. Validation
nDCG@10 selects a checkpoint, including epoch zero.
Float32 norm projection is checked with $10^{-5}$ slack; mathematical bounds
apply with the actual norm, including that numerical tolerance.

For fixed $d,N,m$, the payload size is
\begin{equation}\label{eq:bytes}
 B_{\rm PQ}=4dL+Nm,\qquad B_{\rm adapter}=4dL+Nm+8dr.
\end{equation}
Thus replacing $L$ by $L-2r$ pays for both float32 factors exactly:
\[
 4dL+Nm=4d(L-2r)+Nm+8dr.
\]
Equation~\eqref{eq:bytes} counts the product-code storage of~\cite{PQ}
and adds the factors in~\eqref{eq:calibration}; this storage identity is
derived here for the chosen representation.
Our fixed-budget comparison is PQ64$\times$64 versus
PQ64$\times$48 plus $r=8$ factors. Both use exactly \SciBytes{} bytes on
SciFact and \NFBytes{} bytes on NFCorpus. This equality includes the adapter,
not just the document codes. Common document IDs, text payloads, and encoder
weights are excluded from both sides. No original document embeddings are
needed for scoring. The Python reference reconstructs vectors to batch
matrix products; this is not an optimized lookup-table search engine.

Symmetric $UV^\top+VU^\top$ and general $UV^\top$ adapters use identical
parameter counts, training data, norm bounds, and checkpoints. Their maximal
ranks differ, which is disclosed rather than called a rank-matched control.
Unbounded training and the earlier coarse PQ8 codec are retained in the
results so the successful-looking variant cannot conceal failed trials.
The semantic ablations are exploratory because their variants followed an
inspected pilot; the SIFT experiment uses a separately fixed query partition. The finite calibration mechanism is not asserted to be the
long-time limit of~\eqref{eq:model-main}.

\subsection{What skew geometry guarantees}
\begin{proposition}[Conditioning and retrieval-error certificate]\label{prop:certificate}
For real skew $A$ with $\|A\|_{\rm op}\le b$,
\[
 \|q^\top(I+A)\|^2=\|q\|^2+\|q^\top A\|^2,
 \quad\sigma_{\min}(I+A)\ge1,
 \quad\operatorname{cond}_2(I+A)\le\sqrt{1+b^2}.
\]
If $\|x_i-\hat x_i\|\le r_i$ and $\|q\|=1$, then
\[
 |q^\top(I+A)x_i-s_A(q,i)|\le\sqrt{1+b^2}\,r_i.
\]
Consequently a pairwise exact calibrated margin larger than
$\sqrt{1+b^2}(r_i+r_j)$ preserves that pair's order after quantization.
Relative to uncalibrated cosine, the conservative error is instead
$b+\sqrt{1+b^2}r_i$; calibration changes the target ranking.
\end{proposition}
\begin{proof}
The cross term vanishes since $q^\top A q=0$. Apply the resulting norm
identity to arbitrary vectors and use $\|A\|_{\rm op}\le b$ for both singular
value bounds. Cauchy--Schwarz gives the score error. Subtract the two score
errors to obtain the ranking condition. Finally
$|q^\top A x_i|\le b$ for unit $q,x_i$.
\end{proof}
For $b=1/2$, the condition number is at most $\sqrt{5}/2$. This is an
algebraic guarantee; it does not guarantee improved nDCG or certify top-$k$
relevance from unlabeled geometry. Radius certificates for the much coarser
centroid pilot certify no top-10 sets in the observed data, a useful failure
case rather than a reason to suppress the measurement.

\begin{proposition}[When a skew direction can improve a smooth loss]\label{prop:gradient}
Let $\mathcal L(A)$ be differentiable at zero, with Frobenius gradient $G$,
and put $S=(G-G^\top)/2$. Over skew directions $D$ with $\|D\|_F\le1$,
\[
 \min_D\langle G,D\rangle_F=-\|S\|_F.
\]
There is a strict first-order skew descent direction exactly when $S\ne0$.
If the second directional derivative is bounded above by $H$ along
$D=-S/\|S\|_F$, then
$\mathcal L(tD)\le\mathcal L(0)-t\|S\|_F+Ht^2/2$.
\end{proposition}
\begin{proof}
Symmetric matrices are Frobenius-orthogonal to skew matrices, so
$\langle G,D\rangle_F=\langle S,D\rangle_F$. Cauchy--Schwarz gives the
minimum and its attaining direction. Taylor's integral remainder gives the
finite-step bound. The $S=0$ case excludes strict first-order descent only;
it says nothing about possible second-order improvements.
\end{proof}
This is a standard projected-gradient fact, used as a falsifiable mechanism.
It explains why directionality cannot be assumed beneficial for arbitrary
data. It does not establish that the factorized nonconvex optimizer reaches
the best skew matrix, or that skew is better than an unconstrained adapter.

\section{Particle Growth and the Metastable Time Scale}
For two narrow blocks, normalized interaction weights depend on total block
mass rather than the number of identical or nearby particles. This distinction
leads to Theorem~\ref{thm:block-mass}: if the block-mass ratio is
$e^{o(\beta)}$, the limiting center angle is acute, and the skew pairing of
centers vanishes, the first cross-block clock has exponent $1-\cos\phi$,
without a restriction on particle-count growth. Its proof uses a common
weighted projective kernel to obtain subexponential intrablock relaxation,
then a uniform nonlinear contraction and a cross-block Gram comparison.
The block-mass fractions add to one in the final comparison, eliminating
the minimum-individual-mass factor. For exactly collapsed blocks with mass
ratio $e^{\beta a}$, the exponent changes to $[1-\cos\phi-|a|]_+$.
This identifies an explicit boundary of the count-invariant regime.

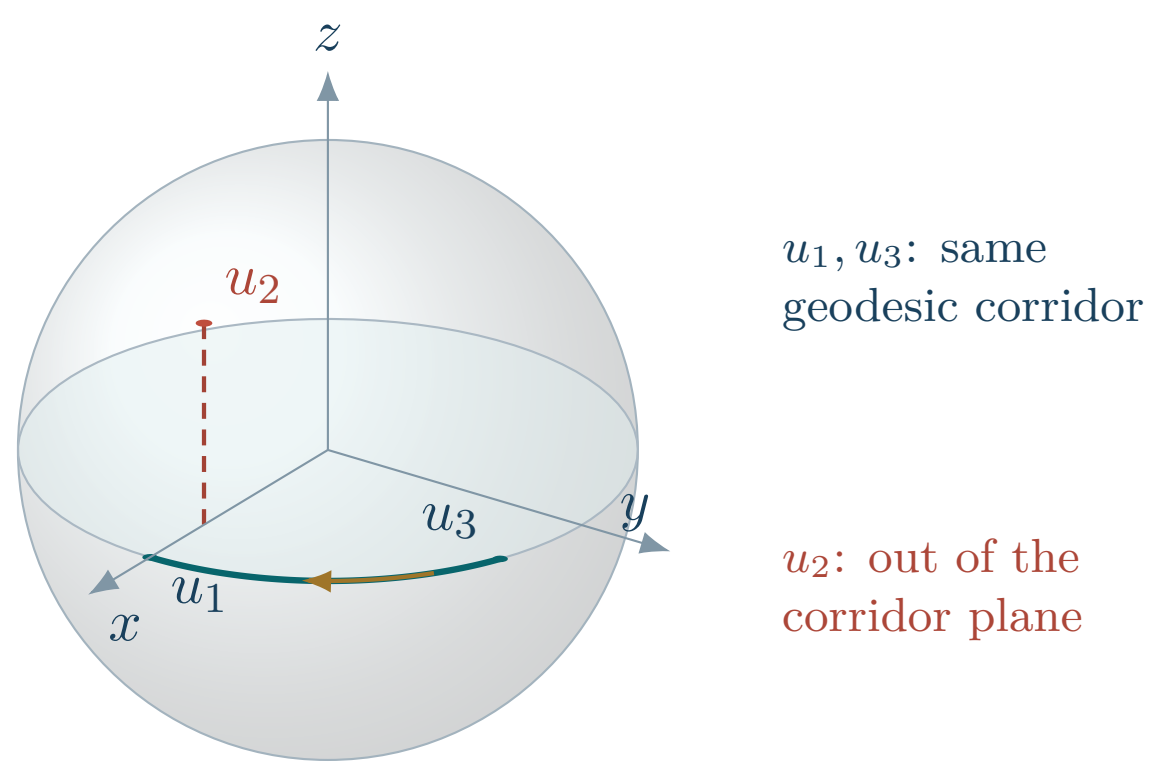
\begin{figure}[H]
\centering
\tdplotsetmaincoords{65}{125}
\begin{tikzpicture}[tdplot_main_coords,scale=2.05,>=Latex]
\begin{scope}[tdplot_screen_coords]
\shade[ball color=soft,opacity=.27] (0,0) circle (1);
\draw[deep!40,thin] (0,0) circle (1);
\end{scope}
\fill[teal!12,opacity=.45] (0,0,0) -- plot[variable=\t,domain=0:360,samples=90] ({cos(\t)},{sin(\t)},0) -- cycle;
\draw[deep!37,thin] plot[variable=\t,domain=0:360,samples=90] ({cos(\t)},{sin(\t)},0);
\draw[teal!85!black,very thick] plot[variable=\t,domain=0:68.7549,samples=60] ({cos(\t)},{sin(\t)},0);
\coordinate (uone) at (1,0,0);
\coordinate (uthree) at ({cos(1.2 r)},{sin(1.2 r)},0);
\coordinate (utwo) at ({cos(.8 r)},0,{sin(.8 r)});
\coordinate (foot) at ({cos(.8 r)},0,0);
\draw[coral!85!black,densely dashed,thick] (utwo)--(foot);
\fill[teal!85!black] (uone) circle (.025) node[below right,deep] {$u_1$};
\fill[teal!85!black] (uthree) circle (.025) node[above left,deep] {$u_3$};
\fill[coral] (utwo) circle (.027) node[above right,coral!90!black] {$u_2$};
\draw[golden!90!black,thick,-{Latex[length=2mm]}] plot[variable=\t,domain=55:30,samples=18] ({cos(\t)},{sin(\t)},0);
\draw[deep!55,-{Latex[length=2mm]}] (0,0,0)--(0,0,1.35) node[above,deep] {$z$};
\draw[deep!55,-{Latex[length=2mm]}] (0,0,0)--(1.35,0,0) node[below right,deep] {$x$};
\draw[deep!55,-{Latex[length=2mm]}] (0,0,0)--(0,1.35,0) node[above left,deep] {$y$};
\begin{scope}[tdplot_screen_coords]
\node[anchor=west,align=left,deep,font=\footnotesize] at (1.4,.53) {$u_1,u_3$: same\\geodesic corridor};
\node[anchor=west,align=left,coral!90!black,font=\footnotesize] at (1.4,-.45) {$u_2$: out of the\\corridor plane};
\end{scope}
\end{tikzpicture}
\caption{Exact three-dimensional point coordinates from~\eqref{eq:example}, shown in an orthographic projection. The blue arc is the shielded great-circle corridor; the dashed line projects $u_2$ onto its plane.}
\label{fig:sphere}
\end{figure}

Figure~\ref{fig:sphere} shows the explicit three-point geometry underlying
the two-clock example: $u_1=(1,0,0)$,
$u_2=(\cos0.8,0,\sin0.8)$, and
$u_3=(\cos1.2,\sin1.2,0)$.
For the planar skew rotation and $\varepsilon=0.6$, the favorable directed
pair is $(3,1)$, although $(1,2)$ is initially closest.
The corridor theorem and Corollary~\ref{cor:full-second} establish the
first-event selection and second-event exponent, respectively.

\section{Experimental Evaluation}
\subsection{Data, separation, and metrics}
We use BEIR~\cite{BEIR} SciFact (5,183 documents) and NFCorpus (3,633
documents), one frozen all-MiniLM-L6-v2 encoder~\cite{MiniLM}, mask-aware
mean pooling, maximum 256 wordpieces, and normalized 384-dimensional
embeddings. Corpus and model revisions and downloaded-file SHA256 values
are recorded. SciFact has 648 fitting, 161 validation, and 300 test queries;
NFCorpus has 2,072 fitting, 518 validation, and 323 test queries. A fixed
20\% query-ID holdout from the official training queries supplies validation;
official test IDs are disjoint. Official development sets are not used.
The corpus is transductively available for unsupervised codebook fitting,
but test qrels are never used for fitting or checkpoint selection.

We report nDCG@10 with exponential relevance gains and Recall@100, averaging
three seeds. Confidence intervals bootstrap whole test queries after
averaging initialization seeds, with 5,000 resamples; these intervals do not
estimate uncertainty across encoders or data sets. No multiple-comparison
adjustment is applied, so the exploratory intervals are descriptive. Encoder
pretraining overlap with BEIR cannot be excluded, and the semantic adapter uses
labels while plain PQ does not. A separate fresh benchmark is needed for a
confirmatory generalization claim. MTEB~\cite{MTEB} motivates evaluation
across tasks rather than treating clustering quality as retrieval quality.

\subsection{Semantic Retrieval and Coarsening Ablations}
\FloatBarrier
\begin{table}[H]\centering\small
\caption{Fixed-byte retrieval comparison. Values are means over three seeds. PQ48 alone uses fewer bytes; all three adapters and PQ64 have identical payload sizes within each dataset.}
\label{tab:fixedbytes}
\begin{tabular}{lrrrr}\toprule
& \multicolumn{2}{c}{SciFact} & \multicolumn{2}{c}{NFCorpus}\\
Method & nDCG@10 & R@100 & nDCG@10 & R@100\\\midrule
PQ64$\times$64 & 0.6124 & 0.9159 & 0.3046 & 0.2925\\
PQ64$\times$48, no adapter & 0.6127 & 0.9131 & 0.3019 & 0.2958\\
PQ48 + bounded skew & 0.6210 & 0.9203 & 0.3049 & 0.3000\\
PQ48 + bounded symmetric & 0.6165 & 0.9196 & 0.3053 & 0.2984\\
PQ48 + bounded general & 0.6222 & 0.9140 & 0.3057 & 0.2989\\
\midrule
Uncompressed float32 & 0.6451 & 0.9250 & 0.3173 & 0.3115\\
\bottomrule\end{tabular}\end{table}
On SciFact, the skew-minus-byte-matched-PQ nDCG difference is +0.0086, with a query-bootstrap 95\% interval [-0.0038, +0.0207].
On NFCorpus, the skew-minus-byte-matched-PQ nDCG difference is +0.0003, with a query-bootstrap 95\% interval [-0.0057, +0.0061].
The SciFact point estimate is positive, but neither fixed-byte confidence interval excludes zero. The general adapter is competitive, and the matched adapter controls do not establish a unique benefit of skew structure in this fine-codec regime. The ancillary records include the coarse PQ8 codec and unbounded-adapter ablations.
The float32 document-vector payload is reduced by a factor of 18.51 on SciFact and 16.87 on NFCorpus. These ratios exclude common encoder, IDs, and text storage and do not describe end-to-end system memory.

\begin{table}[H]\centering\small
\caption{Coarsening ablation on SciFact: nDCG@10 by prototype budget and local perturbation stability.}
\label{tab:coarsening}
\begin{tabular}{lrrrrr}\toprule
& \multicolumn{3}{c}{Prototype budget $K$} & Noise & Neighbor\\
Method & 128 & 256 & 512 & amplification & Jaccard\\\midrule
Spherical clustering & 0.0774 & 0.1350 & 0.1954 & 1.000 & 0.990\\
Reciprocal flow & 0.0866 & 0.1338 & 0.1980 & 0.812 & 0.991\\
Directed flow & 0.0813 & 0.1333 & 0.1950 & 0.811 & 0.991\\
Coarse PQ8 reference & \multicolumn{3}{c}{0.3713 (no prototype budget)} & --- & ---\\
\bottomrule\end{tabular}\end{table}

\subsection{Joint Particle-Count and Temperature Scaling}\label{sec:growth-experiment}
The growing-system theorem predicts a count-independent exponent for balanced
near-collision blocks, even if $N_\beta$ is exponential in $\beta$. We integrate
the full smooth particle ODE with distinct, noncollapsed initial points and
$N_\beta=\lceil e^{\gamma\beta}\rceil$ for $\gamma\in\{0.15,0.25\}$.
The center angle is $0.8$, the threshold is $0.3$, and the skew operator is
nonzero with zero restriction to the center plane. Initial off-plane
perturbations have radius $0.025/\sqrt\beta$. Each block has total mass one;
no forced collisions or quotient replacement are used in this experiment.
A second regime gives individual log masses a range $0.2\beta^2$ within
each block before normalizing each block to total mass one. It violates the
older minimum-individual-mass scaling condition while preserving the
block-mass hypothesis of Theorem~\ref{thm:block-mass}.
\begin{table}[H]\centering\small
\caption{Full-flow joint scaling, $G=1-\cos(0.8)=0.30329329$. Total block masses are balanced. Extreme intrablock weights have logarithmic range $0.2\beta^2$.}
\label{tab:growth}
\begin{tabular}{lrrrrr}\toprule
Weights & $\gamma$ & $\beta$ & $N_\beta$ & $T_\beta$ & $\log(\beta T_\beta)/\beta$\\\midrule
Uniform & 0.15 & 8 & 4 & 2.3634 & 0.36744\\
Uniform & 0.15 & 12 & 7 & 4.4670 & 0.33180\\
Uniform & 0.15 & 16 & 12 & 10.4881 & 0.32018\\
Uniform & 0.15 & 20 & 21 & 26.3024 & 0.31327\\
Uniform & 0.15 & 24 & 37 & 68.7424 & 0.30868\\
Uniform & 0.25 & 8 & 8 & 2.3940 & 0.36905\\
Uniform & 0.25 & 12 & 21 & 4.7178 & 0.33635\\
Uniform & 0.25 & 16 & 55 & 10.3133 & 0.31913\\
Uniform & 0.25 & 20 & 149 & 25.5085 & 0.31174\\
Uniform & 0.25 & 24 & 404 & 68.7768 & 0.30871\\
Extreme & 0.25 & 8 & 8 & 2.3905 & 0.36887\\
Extreme & 0.25 & 12 & 21 & 4.4808 & 0.33206\\
Extreme & 0.25 & 16 & 55 & 10.0135 & 0.31728\\
Extreme & 0.25 & 20 & 149 & 23.7073 & 0.30808\\
Extreme & 0.25 & 24 & 404 & 68.9838 & 0.30883\\
\bottomrule\end{tabular}\end{table}

The diagnostic $\log(\beta T_\beta)/\beta$ removes the leading logarithmic
endpoint factor to expose finite-temperature convergence; the theorem itself
concerns $\log T_\beta/\beta$. These finite computations support the stated
joint-limit regime and do not prove it. The exact-block mass-imbalance check
separately distinguishes the limiting rates $1-\cos(0.8)$ and
$1-\cos(0.8)-0.1$ when the block-mass ratio grows as $e^{0.1\beta}$.

The Gaussian perturbation test adds noise of expected norm approximately
0.01 to corpus embeddings, renormalizes, and measures output displacement
amplification and top-10 neighbor overlap on a deterministic 256-point
sample. It is a specified local stress test, not evidence for stability
under domain shift, different encoders, or adversarial perturbations. The
reciprocal and directed flows show similar attenuation; a small measured
attenuation difference does not establish a directionality advantage.

\subsection{Million-Vector Visual Retrieval}\label{sec:large}

We use all 1,000,000 SIFT-128 database descriptors from the public ANN-Benchmarks dataset~\cite{ANNData}. Each descriptor is normalized to unit length. We recompute exact cosine top-100 ground truth using \texttt{IndexFlatIP}; the dataset's supplied Euclidean neighbors are not reused. The 10,000 queries are split by fixed order into 2,000 fitting, 2,000 validation, and 6,000 test queries. No language model or descriptor fine-tuning is used in this experiment.

Each FAISS~\cite{FAISS} index uses 16 eight-bit PQ subspaces, trained on a fixed 65,536-vector corpus sample. OPQ~\cite{OPQ} uses 15 outer iterations before a separately trained PQ codebook. Three PQ initialization seeds are evaluated. Calibration uses only fitting queries and the same corpus sample; validation nearest-neighbor hit@10 selects $\lambda\in\{10^{-6},10^{-5},10^{-4}\}$. Corrections have spectral norm at most 0.5 and rank at most 16. The general rank-16 control stores twice as many factor entries as the canonical skew representation; the table includes that actual storage difference.

\begin{table}[H]\centering\small

\caption{Normalized SIFT1M: exact-nearest-neighbor hit@10/100, top-10 set recall, persistent serialized index plus correction size, and measured query time. Means over three codebook seeds.}

\label{tab:sift}

\begin{tabular}{lrrrrr}\toprule

Method & Hit@10 & Hit@100 & Set R@10 & MiB & ms/query\\\midrule

PQ & 0.4189 & 0.7752 & 0.2258 & 15.384 & 0.729\\

PQ + skew & 0.4221 & 0.7814 & 0.2283 & 15.392 & 0.730\\

PQ + symmetric & 0.4246 & 0.7843 & 0.2302 & 15.392 & 0.729\\

PQ + general & 0.4257 & 0.7838 & 0.2293 & 15.399 & 0.729\\

OPQ & 0.3477 & 0.7215 & 0.1895 & 15.446 & 0.741\\

OPQ + skew & 0.3558 & 0.7303 & 0.1935 & 15.454 & 0.737\\

OPQ + symmetric & 0.3539 & 0.7301 & 0.1934 & 15.454 & 0.735\\

OPQ + general & 0.3709 & 0.7465 & 0.2012 & 15.462 & 0.729\\

\bottomrule\end{tabular}\end{table}

Hit@$k$ is the fraction of test queries whose exact nearest neighbor appears among the first $k$ returned documents. Set recall@10 is the fraction of the exact top-10 set recovered. Search uses exhaustive compressed lookup-table scans, not reconstructed full vectors. Timings include the OPQ query rotation and factored correction, exclude metric evaluation, use four FAISS threads, and take the median of three full-test searches after warm-up. They are amortized batch-query times on the recorded host, not single-request tail latency.

For PQ, the skew-minus-base hit@10 difference is +0.0031, with query-bootstrap 95\% interval [+0.0008, +0.0053].

For OPQ, the skew-minus-base hit@10 difference is +0.0081, with query-bootstrap 95\% interval [+0.0059, +0.0104].

These intervals average seeds before resampling test queries and do not quantify between-dataset uncertainty. Neither a smaller calibration loss nor a positive point estimate is treated as a universal retrieval advantage. The score-calibration theorem certifies its convex distortion objective; measured neighbor recovery remains a separate empirical quantity.

\paragraph{Native-distance controls.} Because the exact descriptors are unit length, squared-L2 and cosine exact rankings agree. Quantized reconstructions have varying norms, so their compressed L2 and inner-product rankings can differ. The following native-distance controls are therefore required before interpreting a calibration gain.

\begin{table}[H]\centering\small

\caption{Strong native-L2 controls and calibrated hard-negative margins on SIFT1M. Query correction is included in timing; all values average three codebook seeds.}

\begin{tabular}{lrrrr}\toprule

Method & Hit@10 & Hit@100 & MiB & ms/query\\\midrule

PQ-L2 & 0.8926 & 0.9981 & 15.384 & 0.731\\

PQ-L2 + score skew & 0.8905 & 0.9974 & 15.392 & 0.737\\

OPQ-L2 & 0.8357 & 0.9941 & 15.446 & 0.758\\

OPQ-L2 + score skew & 0.8288 & 0.9928 & 15.454 & 0.753\\

PQ-L2 + margin skew & 0.8948 & 0.9981 & 15.392 & 0.745\\

PQ-L2 + margin symmetric & 0.8996 & 0.9986 & 15.392 & 0.747\\

PQ-L2 + margin general & 0.8991 & 0.9987 & 15.399 & 0.733\\

\bottomrule\end{tabular}\end{table}

The margin experiment uses 16 compressed-search hard negatives per fitting query and the norm-aware target in Section~\ref{sec:margin}. Validation chooses among ridge penalties $10^{-5},10^{-4},10^{-3}$; the smaller initial penalty grid failed the solver stopping criterion and was replaced by a diagonally preconditioned solve before this final evaluation. All final residuals and validation grids are retained. Native L2 changes recovery much more than the skew score correction. Margin calibration improves its target alignment, while symmetric and general margin controls remain stronger in this experiment; no unique empirical advantage of skew is inferred.

The margin-skew minus native PQ-L2 hit@10 difference is +0.0022, with query-bootstrap interval [+0.0002, +0.0043].

\begin{table}[H]\centering\small

\caption{Rank sensitivity of the score-distortion skew correction for native PQ-L2, seed 0. The ridge penalty is fixed to its validation-selected value.}

\begin{tabular}{rrrrr}\toprule

Rank & Hit@10 & Hit@100 & Correction bytes & Distortion objective\\\midrule

2 & 0.8913 & 0.9978 & 1024 & 2.765e-04\\

4 & 0.8910 & 0.9978 & 2048 & 2.755e-04\\

8 & 0.8913 & 0.9978 & 4096 & 2.753e-04\\

16 & 0.8912 & 0.9978 & 8192 & 2.752e-04\\

32 & 0.8917 & 0.9978 & 16384 & 2.751e-04\\

\bottomrule\end{tabular}\end{table}

\subsection{Ten-Million-Vector Search and Memory Compression}\label{sec:engineering}

We ingest the first 10,000,000 real BIGANN SIFT descriptors from the public billion-scale benchmark archive, without duplicating the one-million-vector dataset. The source prefix is obtained by an exact HTTP byte range, its length and SHA256 are recorded, and the descriptors are normalized in streaming blocks. Exact cosine top-100 neighbors are recomputed for all 10,000 public queries. The same fixed 2,000/2,000/6,000 fitting/validation/test partition is used. The public BIGANN query file has no exact descriptor overlap with the SIFT1M query file in this run.

The engineered index is native FAISS~\cite{FAISS} IVF4096,PQ16x8 with squared-L2 scoring. It is trained on 262,144 sampled corpus descriptors and ingested in blocks of 200,000. Search uses an Apple M4 Pro host with 48 GiB RAM, macOS 26.3, and four CPU threads. We report a single built index and a probe-budget sweep, rather than treating probe settings as independent seeds. Serialized size includes vector IDs, inverted lists, codebooks, and coarse centers; it is a persistent payload measurement, not process peak RAM.

\begin{table}[H]\centering\small

\caption{BIGANN10M IVF-PQ search: 6,000 held-out queries. Median of three timed batch searches, four CPU threads.}

\begin{tabular}{rrrrr}\toprule

Probes & Approx. scanned vectors & Hit@10 & Hit@100 & ms/query\\\midrule

4 & 9,766 & 0.6722 & 0.7080 & 0.0233\\

16 & 39,062 & 0.8402 & 0.9202 & 0.0568\\

64 & 156,250 & 0.8827 & 0.9862 & 0.1794\\

256 & 625,000 & 0.8862 & 0.9963 & 0.6159\\

\bottomrule\end{tabular}\end{table}

The serialized index occupies 231.04 MiB, compared with 4.77 GiB for raw float32 vectors. Training takes 17.21 seconds and streaming ingestion 66.30 seconds on the recorded host. Ground-truth construction is excluded from both build timings.

The end-to-end skew margin control is also run through IVF routing and candidate scoring. Its validation protocol includes the identity update; validation selects that identity here. Accordingly its test rankings are unchanged and no ten-million-vector calibration gain is claimed. This guards against attributing index-parameter effects to a learned correction.

\paragraph{Mass-centered attention memory.} To test the finite-compression theorem directly, we aggregate each of the 4,096 IVF cells by its exact count and the arithmetic mean of its original normalized keys. A streamed pass computes exact attention over all 10 million keys for 1,024 held-out queries at three inverse temperatures, with a fixed rank-two skew interaction of strength 0.3. The comparison retains identical cell masses and contrasts the exact barycenters, normalized barycenters, and the original sample-trained coarse centroids. It therefore isolates centering/normalization effects rather than changing the partition.

\begin{table}[H]\centering\small

\caption{Ten-million-key attention memory: mean Euclidean output error against exact streamed attention. All three methods retain 4,096 centers and the same cell masses.}

\begin{tabular}{rrrr}\toprule

$\beta$ & Mass-centered & Normalized centers & Trained coarse centers\\\midrule

1 & 0.003735 & 0.095486 & 0.003869\\

8 & 0.028480 & 0.096562 & 0.026600\\

32 & 0.095027 & 0.117941 & 0.091391\\

\bottomrule\end{tabular}\end{table}

The center-and-count payload is 2.031 MiB, versus 5.12 GB of original key vectors. This compresses an attention-memory readout, not document identities or exact nearest-neighbor rankings. The measured error increases at high inverse temperature, consistently with the explicit temperature factors in Theorem~\ref{thm:centered}; the small-radius theorem is not interpreted as a uniformly small error bound for arbitrary IVF cells.

\FloatBarrier
\subsection{Computational Complexity and Evaluation Scope}
The dense coarsening reference requires $O(TN^2d)$ arithmetic and $O(N^2)$
attention storage. The deployed correction is evaluated through compressed
FAISS indices, using $O(dr)$ query-side work for a rank-$2r$ factorization.
Section~\ref{sec:large} measures complete lookup-table search separately from
metric evaluation, with native PQ and optimized product quantization (OPQ)
controls. Quality, persistent index size, and measured query time are reported
jointly; reconstruction error alone is not used as a retrieval claim.

\section{Second-Order Mass-Centered Attention Compression}\label{sec:second-order}
Exact mass aggregation becomes quantitatively stronger when the compression
center cancels the first moment. In a block $I_a$, put
$m_a=\sum_{j\in I_a}m_j$, $w_{aj}=m_j/m_a$ and
$c_a=\sum_jw_{aj}u_j$, without normalizing $c_a$. Assume
$\|u_j-c_a\|\le r_a\le r$ and $\|u_j\|=1$. Let $\bar p_a(x)$ be the softmax
using masses $m_a$ and scores $\beta x^\top M c_a$ for a unit query $x$.
The factor $m_a$ is the mass correction already used for proportional
attention in token merging~\cite[Section~3, Eq.~(1)]{ToMe}; for unit item masses it
adds $\log|I_a|$ to the cluster logit. Size and dispersion corrections in
two-level softmax are analyzed in~\cite[Sections~3--4]{TwoLevel}.
The theorem below controls the vector output and the additional error
caused by normalizing the barycenter.

\begin{theorem}[Centered log-partition and attention error]\label{thm:centered}
Write $K=\|M\|_{\rm op}$ and $h=\beta Kr$. The exact block probabilities
are obtained from $\bar p_a$ by adding logit corrections $\delta_a$ with
\[
 0\le\delta_a=\log\sum_{j\in I_a}w_{aj}
       e^{\beta x^\top M(u_j-c_a)}\le h^2/2.
\]
If $\mu(x)$ is the full attention output and
$\bar\mu(x)=\sum_a\bar p_a(x)c_a$, then
\begin{equation}\label{eq:centered-error}
 \|\mu(x)-\bar\mu(x)\|
 \le\beta K r^2+\min\{2,e^{h^2/2}-1\}.
\end{equation}
For fixed $\beta,K$, this is $O(r^2)$ rather than the uncentered
$O(r)$ bound. The order $r^2$ cannot be uniformly improved.
\end{theorem}
\begin{proof}
In each block the residuals $z_j=u_j-c_a$ have weighted mean zero. The scalar
$Y_j=\beta x^\top Mz_j$ lies in $[-h,h]$ and has mean zero. Jensen's
inequality gives the lower bound. Hoeffding's bounded-variable
log-moment inequality~\cite{Hoeffding}, with range length $2h$,
gives $\log\mathbb E e^Y\le(2h)^2/8=h^2/2$.
Factoring the exact block numerator as
$m_a e^{\beta x^\top Mc_a}e^{\delta_a}$ proves the probability formula.
If $D=h^2/2$, normalization multiplies each $\bar p_a$ by a ratio in
$[e^{-D},e^D]$, so the probability $\ell^1$ error is at most
$\min(2,e^D-1)$.

It remains to compare the within-block, tilted mean to $c_a$. Define
\[
 \nu_a(t)=\frac{\sum_j w_{aj}e^{t x^\top M u_j}u_j}
                    {\sum_jw_{aj}e^{t x^\top M u_j}}.
\]
Direct differentiation gives
$\nu_a'(t)=\operatorname{Cov}_{a,t}(u)M^\top x$.
The covariance is positive semidefinite and
\[
 \|\operatorname{Cov}_{a,t}(u)\|_{\rm op}
 \le\operatorname{tr}\operatorname{Cov}_{a,t}(u)
 \le\mathbb E_{a,t}\|u-c_a\|^2\le r^2.
\]
Hence $\|\nu_a(\beta)-c_a\|\le\beta Kr^2$. Sum this estimate with the exact
block probabilities and then bound the probability replacement using
$\|c_a\|\le1$. This proves~\eqref{eq:centered-error}.

For sharpness take one equal-mass block with
$u_\pm=(\cos\theta,\pm\sin\theta)$, $M=I$, and $x=(0,1)$.
The center is $(\cos\theta,0)$, $r=\sin\theta$, and the exact second output
coordinate is $\sin\theta\tanh(\beta\sin\theta)$.
For fixed $\beta>0$ its ratio to $r^2$ tends to $\beta$, so a uniform
$o(r^2)$ output bound is impossible. Also
$\delta=\log\cosh(\beta r)=\beta^2r^2/2+O(r^4)$, attaining the second-order
log-partition coefficient asymptotically.
\end{proof}

If a spherical representative is required, use $v_a=c_a/\|c_a\|$ whenever
$r<1$. The unit-key identity
$1-\|c_a\|^2=\sum_jw_{aj}\|u_j-c_a\|^2\le r^2$ implies
$\|v_a-c_a\|=1-\|c_a\|\le r^2$. Applying the logit perturbation estimate
once more adds at most $r^2+\min(2,e^{2\beta Kr^2}-1)$ to
\eqref{eq:centered-error}. This explicitly accounts for normalization rather
than treating a Euclidean barycenter as a unit vector.

The aggregation estimate controls attention-output error, whereas the calibrated
quadratic controls retrieval-score distortion. Their joint use is explicit:
a partition first fixes masses and representatives; its centered residuals
then enter $D$ and $E$ in the calibration objective. The residual certificate
quantifies the additional optimization error. These quantities can therefore
be measured separately instead of treating small cluster radius as an
unqualified guarantee of ranking preservation.

\section{Verification and Reproducibility}
The additional equation audit checks 60 independently assembled skew-basis systems (maximum solution difference 9.76e-13), automatic derivatives and centered finite differences (maximum directional error 5.47e-13), and 240 centered-attention configurations. A directed-rounding interval residual enclosure is included for a small exact-float input system. It bounds the optimization gap for that instance, rather than claiming interval certification of the entire large benchmark. A further 300 heterogeneous-mass configurations test the projective common-kernel diameter inequality used by the growing-block theorem; the maximum projective differential-identity error is 6.66e-16, with no contraction-bound violations. These finite tests complement the analytic argument.

The analytic ODE theorems have handwritten proofs and are not fully formalized.
The checked file \texttt{formal/Algebra.v} proves nine kernels: finite-sum
Gram product algebra; duplicate-mass summation; planar cycle-area algebra;
skew-ratio cancellation under $s^2=1-c^2$; the transverse-contraction
constant inequality; a pairwise ranking-gap inequality; the algebraic
relative-clock comparison; cap-minimum positivity; and the projective
coordinate cancellation. Rocq 9.1.1 compilation and the independent
\texttt{coqchk} library checker both succeed. No project-defined axioms or
admitted proofs are used. The assumption log explicitly lists standard
classical-real and functional-extensionality dependencies. We do not call
these proofs axiom-free.

Lean/mathlib is a viable alternative for a larger analytic development,
with axiom auditing via \texttt{\#print axioms}~\cite{Lean}; Rocq's small
kernel serves the corresponding check here~\cite{Rocq}. Neither system
automatically certifies a prose theorem by recognizing its displayed
equations. The formal statement must faithfully encode the hypotheses and
analysis. SymPy cancellation, randomized differential identities, finite
differences for the rotating spectrum, a second ODE integration method,
and counterexample tests supply complementary evidence but not substitutes
for that formalization. The audit runner fails on missing evidence, admitted
proofs, broken byte budgets, test-label overlap, or numerical identity
failures; unresolved conjectures remain explicitly unresolved.

\section{Discussion}
The spherical flow and compressed scorer use the same skew bilinear geometry,
but their guarantees concern different objects: metastable event times,
attention output approximation, and query--document ranking distortion.
The second-order aggregation bound and the certified calibration objective
provide explicit links between these objects without identifying a finite
query update with an infinite-time coarsening limit. The empirical comparisons
are interpreted within the tested descriptors, index budgets, and query splits.
The full-flow prefactor requires stronger synchronization than its logarithmic
exponent and is not inferred from a numerical fit.

\clearpage
\appendix
\FloatBarrier
\section{Closed Gram Geometry and Supporting Analysis}\label{sec:gram-geometry}
Fix integers $n,d\ge2$, masses $m_i>0$, $\beta>0$, a real skew-symmetric $A\in\R^{d\times d}$, and $\eps\in\R$. Set $M=I_d+\eps A$. For $U=(u_1,\ldots,u_n)$ with $u_i\in\sphere^{d-1}$, define
\begin{equation}\label{eq:model}
\dot u_i=(I_d-u_iu_i^\T)\sum_{j=1}^n p_{ij}(U)u_j,\quad
p_{ij}(U)=\frac{m_j\exp(\beta u_i^\T M u_j)}{\sum_{\ell=1}^n m_\ell\exp(\beta u_i^\T M u_\ell)}.
\end{equation}
All solutions exist uniquely for all $t\ge0$ on the compact manifold $(\sphere^{d-1})^n$.

The \emph{directed barrier} is
\begin{equation}\label{eq:barrier}
g_{ij}(U)=1-u_i^\T M u_j=1-C_{ij}-\eps H_{ij},\qquad
\frac{p_{ij}}{p_{ii}}=\frac{m_j}{m_i}e^{-\beta g_{ij}},
\end{equation}
where $C_{ij}=u_i^\T u_j$ and $H_{ij}=u_i^\T Au_j$.  In particular $C=C^\T$, $H=-H^\T$, and
\begin{equation}\label{eq:asymmetry}
g_{ij}-g_{ji}=-2\eps H_{ij}.
\end{equation}
Although both transition probabilities are positive, their scales need not be comparable.

\subsection{Simultaneous congruence of two Gram matrices}
Write $U\in\R^{n\times d}$ for the matrix with rows $u_i^\T$.  Set
\begin{equation}
P=(p_{ij}),\qquad b_i=\sum_jp_{ij}C_{ij},\qquad B=P-\operatorname{diag}(b_1,\ldots,b_n).
\end{equation}
Then $\dot U=BU$ exactly. This simple matrix identity leads to a useful invariant geometry.
\begin{theorem}[Closed Gram dynamics and preserved pencil]\label{thm:gram}
The pair $(C,H)=(UU^\T,UAU^\T)$ obeys the closed system
\begin{equation}\label{eq:gram}
\boxed{\quad \dot C=BC+CB^\T,\qquad \dot H=BH+HB^\T.\quad}
\end{equation}
The coefficient matrix $B$ is determined by $C,H$ alone through
\begin{equation}
P_{ij}=\frac{m_j e^{\beta(C_{ij}+\eps H_{ij})}}{\sum_km_k e^{\beta(C_{ik}+\eps H_{ik})}}.
\end{equation}
For every finite $t$ there is an invertible $\Phi(t)$ with
\begin{equation}\label{eq:congruence}
C(t)=\Phi(t)C(0)\Phi(t)^\T,\quad
H(t)=\Phi(t)H(0)\Phi(t)^\T.
\end{equation}
Consequently $\operatorname{rank}U(t)$, $\operatorname{rank}H(t)$, and the common row span of $U(t)$ are preserved at finite times. If $C(0)$ is nonsingular, the eigenvalues of $C(t)^{-1}H(t)$ are constant in time.
\end{theorem}
\begin{proof}
Differentiate $UU^\T$ and $UAU^\T$, using $\dot U=BU$, to obtain~\eqref{eq:gram}.  The fundamental solution $\dot\Phi=B(t)\Phi$, $\Phi(0)=I_n$, is invertible at finite times and satisfies $U(t)=\Phi(t)U(0)$. Hence~\eqref{eq:congruence}.  For invertible $C$,
$C(t)^{-1}H(t)=\Phi(t)^{-\T}[C(0)^{-1}H(0)]\Phi(t)^\T$,
which is a similarity transform.  The remaining assertions follow from invertibility of $\Phi$.
\end{proof}
\begin{remark}
For $n>d$, $C$ is necessarily singular and the statement concerning $C^{-1}H$ is unavailable.  The simultaneous congruence~\eqref{eq:congruence}, and hence the algebraic matrix pencil $H-\lambda C$, nevertheless remain meaningful. This invariant is structural, not a general formula for merger order.
\end{remark}

\subsection{Cyclic affinity equals a directed geometric area}
Define the logarithmic affinity of a three-cycle by
\begin{equation}
\mathscr A_{abc}=\log\frac{p_{ab}p_{bc}p_{ca}}{p_{ba}p_{cb}p_{ac}}.
\end{equation}
All mass and normalization factors cancel:
\begin{equation}\label{eq:affinity}
\boxed{\mathscr A_{abc}
=2\beta\eps\bigl(H_{ab}+H_{bc}+H_{ca}\bigr)
=2\beta\eps\,(u_a-u_c)^\T A(u_b-u_c).}
\end{equation}
If $d=2$ and $A=J=\begin{psmallmatrix}0&-1\\1&0\end{psmallmatrix}$, the last expression is $-4\beta\eps$ times the signed Euclidean area of the triangle $(u_a,u_b,u_c)$.  Thus the instantaneous Kolmogorov cycle criterion becomes an \emph{oriented area condition}.  It concerns reversibility of the instantaneous positive Markov matrix $P(U)$, not time reversibility of the nonlinear ODE.

\FloatBarrier
\section{A global consensus region without a gradient structure}
Despite the lack of reciprocal weights, an acute configuration is globally attracted to consensus. This is a quantitative specialization of prior cone-collapse results, not a new consensus mechanism: the proof of \cite[Lemma 4.2]{Perspective} explicitly extends to general query--key matrices. Lemma~\ref{lem:projective} gives the standard projective-coordinate extension to a whole open hemisphere.
\begin{theorem}[Acute-cone consensus]\label{thm:acute}
If $u_i(0)^\T u_j(0)>0$ for all $i,j$, then every pair satisfies
$u_i(t)^\T u_j(t)\to1$ as $t\to\infty$. More precisely, letting
\begin{equation}
\rho(t)=\min_{i,j}u_i(t)^\T u_j(t),\qquad
\kappa_\beta=\frac{m_{\min}}{\sum_jm_j}\exp(-2\beta\|M\|_{\rm op}),
\end{equation}
one has $\rho(t)\ge\rho(0)>0$ and, at almost every $t$,
\begin{equation}\label{eq:consensus}
\dot\rho(t)\ge2\kappa_\beta\bigl(1-\rho(t)^2\bigr).
\end{equation}
In particular, $1-\rho(t)$ decays exponentially in $t$ for each fixed $\beta$.
\end{theorem}
\begin{proof}
For $c_{ij}=u_i^\T u_j$ direct differentiation gives
\begin{equation}\label{eq:pairgram}
\dot c_{ij}=\sum_\ell p_{i\ell}(c_{\ell j}-c_{i\ell}c_{ij})
+\sum_\ell p_{j\ell}(c_{i\ell}-c_{j\ell}c_{ij}).
\end{equation}
Choose an active minimizing off-diagonal pair $(i,j)$. If $\rho\ge0$, then
$c_{\ell j}\ge\rho$ and $c_{i\ell}\le1$, so each term in the first sum is nonnegative; the second sum is handled identically.  Retaining only $\ell=j$ in the first sum and $\ell=i$ in the second yields
$\dot c_{ij}\ge(p_{ij}+p_{ji})(1-\rho^2)$. For every configuration, $p_{ij}\ge\kappa_\beta$. The finite minimum of smooth functions is absolutely continuous and satisfies the lower differential estimate almost everywhere. Scalar comparison gives~\eqref{eq:consensus} and the conclusion.
\end{proof}
\begin{remark}
The convergence theorem does not give a sharp \emph{merger order}. It also does not assert consensus for arbitrary non-acute configurations, as the rotating triangle in \cref{sec:rotation} demonstrates.
\end{remark}

\FloatBarrier
\section{The exactly integrable two-body subsystem}\label{sec:two}
For $n=2$, set $c=u_1^\T u_2$ and $h=u_1^\T Au_2$.  Writing $q_{12}=p_{12}$ and $q_{21}=p_{21}$, the equations reduce to
\begin{equation}\label{eq:two}
\dot c=(q_{12}+q_{21})(1-c^2),\qquad
\dot h=-(q_{12}+q_{21})ch.
\end{equation}
Thus the signed ratio
\begin{equation}\label{eq:alpha}
\boxed{\alpha=\frac{h(t)}{\sqrt{1-c(t)^2}}=\frac{h(0)}{\sqrt{1-c(0)^2}}}
\end{equation}
is conserved whenever $-1<c<1$. The stronger statement with the \emph{sign} of $\alpha$ follows from~\eqref{eq:two}; merely conserving its square would lose the orientation.

For $c_0\in(-1,1)$, define $q_{12}(c)$ and $q_{21}(c)$ by substituting $h=\alpha\sqrt{1-c^2}$ into
\begin{equation}\label{eq:q}
q_{12}=\frac{1}{1+(m_1/m_2)e^{\beta(1-c-\eps h)}},\qquad
q_{21}=\frac{1}{1+(m_2/m_1)e^{\beta(1-c+\eps h)}}.
\end{equation}
The exact hitting time of $c_*\in(c_0,1)$ is
\begin{equation}\label{eq:integral}
T_\beta(c_*)=\int_{c_0}^{c_*}\frac{\dd c}{(1-c^2)(q_{12}(c)+q_{21}(c))}.
\end{equation}
\begin{proposition}[Two-body exponential clock]\label{prop:clock}
For fixed masses, $c_0,c_*,A,\eps$,
\begin{equation}\label{eq:twoasym}
\lim_{\beta\to\infty}\frac{\log T_\beta(c_*)}{\beta}
=\bigl[1-c_0-|\eps h_0|\bigr]_+.
\end{equation}
If $B_0=1-c_0-|\eps h_0|>0$, the hitting time is exponentially long with rate $B_0$.
\end{proposition}
\begin{proof}
Uniformly for $c$ in the compact interval $[c_0,c_*]$, the inverse two-sided transition rate has logarithmic exponential rate equal to
$[f(c)]_+$ where
$f(c)=1-c-|\eps\alpha|\sqrt{1-c^2}$.
The function $f$ is convex and $f(1)=0$. Hence its maximum on $[c_0,c_*]$ cannot exceed $\max(f(c_0),0)$; if $f(c_0)>0$ the maximum is attained at $c_0$, and otherwise the positive part vanishes throughout.  Applying the Laplace principle to~\eqref{eq:integral}, whose nonexponential factor is bounded above and below, proves~\eqref{eq:twoasym}.
\end{proof}

\FloatBarrier
\section{Directed corridor selection of a first threshold event}\label{sec:corridor}
A first merger means the first crossing of a prescribed \emph{positive} angular threshold. Exact collision at finite time cannot occur by uniqueness of the smooth flow.
\begin{definition}[Geometric isolation corridor]\label{def:corridor}
Select an ordered pair $(a,b)$ with initial angle $\theta_0\in(0,\pi)$, a threshold $\theta_*\in(0,\theta_0)$, and initial points $v_i=u_i(0)$.
Let $\gamma$ be the shorter oriented great-circle arc from $v_a$ to $v_b$. Define $\cC_{ab}$ as the compact family of configurations in which $u_a,u_b$ lie on $\gamma$ in their original order, satisfy $\theta_*\le\ang(u_a,u_b)\le\theta_0$, and every $u_j$ with $j\notin\{a,b\}$ is fixed at $v_j$.
The pair has a \emph{strictly shielded corridor} if, with
\begin{equation}
B_0=\min\{g_{ab}(U_0),g_{ba}(U_0)\}>0,
\end{equation}
there are constants $\Delta,\sigma>0$ such that throughout $\cC_{ab}$:
\begin{enumerate}[label=(\roman*),leftmargin=*]
\item Every off-diagonal directed edge $(i,j)$, with $i\ne j$, other than $(a,b),(b,a)$ has $g_{ij}\ge B_0+\Delta$.
\item Every unordered pair other than $\{a,b\}$ has angular separation at least $\theta_*+\sigma$.
\end{enumerate}
\end{definition}
The full off-diagonal-edge condition in (i) is deliberately stronger than needed; it makes the proof transparent and includes all possible competitors.

\begin{theorem}[First-merger selection and sharp exponential rate]\label{thm:corridor}
Under \cref{def:corridor}, for all sufficiently large $\beta$, the pair $\{a,b\}$ is the unique first pair to reach angle $\theta_*$. If $\tau_\beta$ denotes this first hitting time, then
\begin{equation}\label{eq:main}
\boxed{\lim_{\beta\to\infty}\frac{\log\tau_\beta}{\beta}=B_0.}
\end{equation}
The theorem holds for arbitrary fixed positive masses and any fixed skew-symmetric $A$.
\end{theorem}
\begin{proof}
We provide the bootstrap details, since a bare comparison of initial barriers is insufficient. Write $K=\|M\|_{\rm op}$. Each barrier is $2K$-Lipschitz in the max norm of configurations. In a fixed sufficiently small tubular neighborhood of $\cC_{ab}$, all external barriers exceed $B_0+3\Delta/4$ and all external angular gaps exceed $\theta_*+\sigma/2$.

The two moving points on the same oriented great circle have $u_a^\T Au_b=\alpha\sin\theta$ with a fixed signed $\alpha$ determined by the plane and its orientation. Thus the favorable internal barrier is
\begin{equation}\label{eq:f}
f(\cos\theta)=1-\cos\theta-|\eps\alpha|\sin\theta.
\end{equation}
As in \cref{prop:clock}, this is a convex function of $\cos\theta$, equals $B_0$ at the original separation, and equals zero at coincident points. It is therefore bounded above by $B_0$ along the corridor. In a sufficiently small tube, $\min(g_{ab},g_{ba})\le B_0+\Delta/8$.

Because $p_{ij}/p_{ii}=(m_j/m_i)e^{-\beta g_{ij}}$, uniform positivity of masses yields the estimates
\begin{equation}\label{eq:dominance}
q(t):=p_{ab}(t)+p_{ba}(t)\ \ge c\,e^{-\beta(B_0+\Delta/8)},\qquad
\eta_\beta:=\max_i\sum_{\substack{j\ne i\\(i,j)\notin\{(a,b),(b,a)\}}}p_{ij}\ \le C e^{-\beta(B_0+3\Delta/4)}.
\end{equation}
More precisely, in the second bound we sum over \emph{edges outside} the chosen two directed edges, including edges emanating from external points. This notational convention will be used below. The internal two-body vector field moves $u_a,u_b$ toward one another along their common great circle with nonnegative scalar speeds $p_{ab}\sin\theta$ and $p_{ba}\sin\theta$. The external field has max norm at most $C\eta_\beta$.

As long as the solution remains in the tube and $\theta\in[\theta_*,\theta_0]$, the angular equation obeys
\begin{equation}
\dot\theta\le-q\sin\theta+C\eta_\beta\le-\tfrac12q\sin\theta<0
\end{equation}
for large $\beta$. The crossing time is consequently at most $C e^{\beta(B_0+\Delta/8)}$. If $E$ is the maximal normal deviation, then $D^+E\le q(1-c)E+C\eta_\beta$. On the fixed interval $[\theta_*,\theta_0]\subset(0,\pi)$, one has $(1-c)/\sin\theta=\tan(\theta/2)\le\tan(\theta_0/2)$. Hence $\int q(1-c)\,dt\le C\int q\sin\theta\,dt=O(1)$ before threshold crossing, including when $c$ is initially negative. Gronwall therefore controls normal amplification uniformly. During this period external points, and the normal deviations of $u_a,u_b$ away from the oriented great circle, move by at most $C e^{-5\beta\Delta/8}$. Tangential components of the internal field point into the arc, while the integrated external error is of the same exponentially small order. A first-exit bootstrap closes the tube assumption. All competing angular gaps stay above $\theta_*$, proving selection.

To sharpen the clock, the distance of the true solution from $\cC_{ab}$ is $O(e^{-5\beta\Delta/8})$. Hence the favorable internal barrier is at most $B_0+O(e^{-5\beta\Delta/8})$ and $q\ge c e^{-\beta B_0}$ for large $\beta$. The angular comparison improves the bound to $\tau_\beta\le C e^{\beta B_0}$.

Conversely, for arbitrary $\nu\in(0,B_0)$ a sufficiently small neighborhood of $U_0$ has every off-diagonal directed barrier at least $B_0-\nu$. Within this neighborhood all particle speeds are at most $C_\nu e^{-\beta(B_0-\nu)}$. A fixed positive displacement is necessary before the angular threshold can be reached, so $\tau_\beta\ge c_\nu e^{\beta(B_0-\nu)}$. Taking logarithms, then $\beta\to\infty$ and $\nu\downarrow0$, gives~\eqref{eq:main}.
\end{proof}
\begin{remark}[Interpretation and scope]
The selection theorem is \emph{not} the statement that the globally smallest initial barrier always wins. It requires a barrier gap throughout an explicit continuum of geometric configurations. It does not claim permanence of a pair after threshold crossing.
\end{remark}

\FloatBarrier
\section{An analytic three-point inversion of Euclidean proximity}\label{sec:example}
Consider $n=d=3$, $m_i=1$, and
\begin{equation}\label{eq:example}
\begin{gathered}
\eps=0.6,\qquad A=\begin{pmatrix}0&-1&0\\1&0&0\\0&0&0\end{pmatrix},\\[4pt]
u_1=(1,0,0),\quad u_2=(\cos0.8,0,\sin0.8),\quad u_3=(\cos1.2,\sin1.2,0).
\end{gathered}
\end{equation}
Angles here and in the figures are measured in radians. The three initial angles are
\begin{equation}
\theta_{12}=0.8,\quad \theta_{13}=1.2,\quad
\theta_{23}=\arccos(\cos0.8\cos1.2)\approx1.3156.
\end{equation}
Thus the closest pair is $(1,2)$, not $(1,3)$. But the favorable edge of $(1,3)$ has barrier
\begin{equation}\label{eq:Bexample}
B_0=g_{31}(U_0)=1-\cos1.2-0.6\sin1.2\approx0.07841879.
\end{equation}
By contrast $g_{12}=g_{21}=1-\cos0.8\approx0.30329329$.

For arbitrary moving points $v$ on the $xy$-arc $0\le\phi\le1.2$, the external scores against the fixed $u_2$ satisfy
\begin{equation}\label{eq:shield}
v^\T M u_2=\cos0.8(\cos\phi+0.6\sin\phi)
\le\cos0.8\sqrt{1+0.6^2},
\end{equation}
while $u_2^\T Mv=\cos0.8(\cos\phi-0.6\sin\phi)\le\cos0.8$. Consequently
\begin{equation}\label{eq:Gexample}
g_{\rm external}\ge1-\cos0.8\sqrt{1+0.6^2}\approx0.187507>B_0.
\end{equation}
All external angular gaps are at least $0.8$. With $\theta_*=0.3$, \cref{thm:corridor} applies and guarantees
\begin{equation}
\tau_{13}(0.3)<\min\{\tau_{12}(0.3),\tau_{23}(0.3)\}
\quad\text{for all sufficiently large }\beta,
\end{equation}
with $\lim_{\beta\to\infty}\beta^{-1}\log\tau_{13}(0.3)=B_0$.

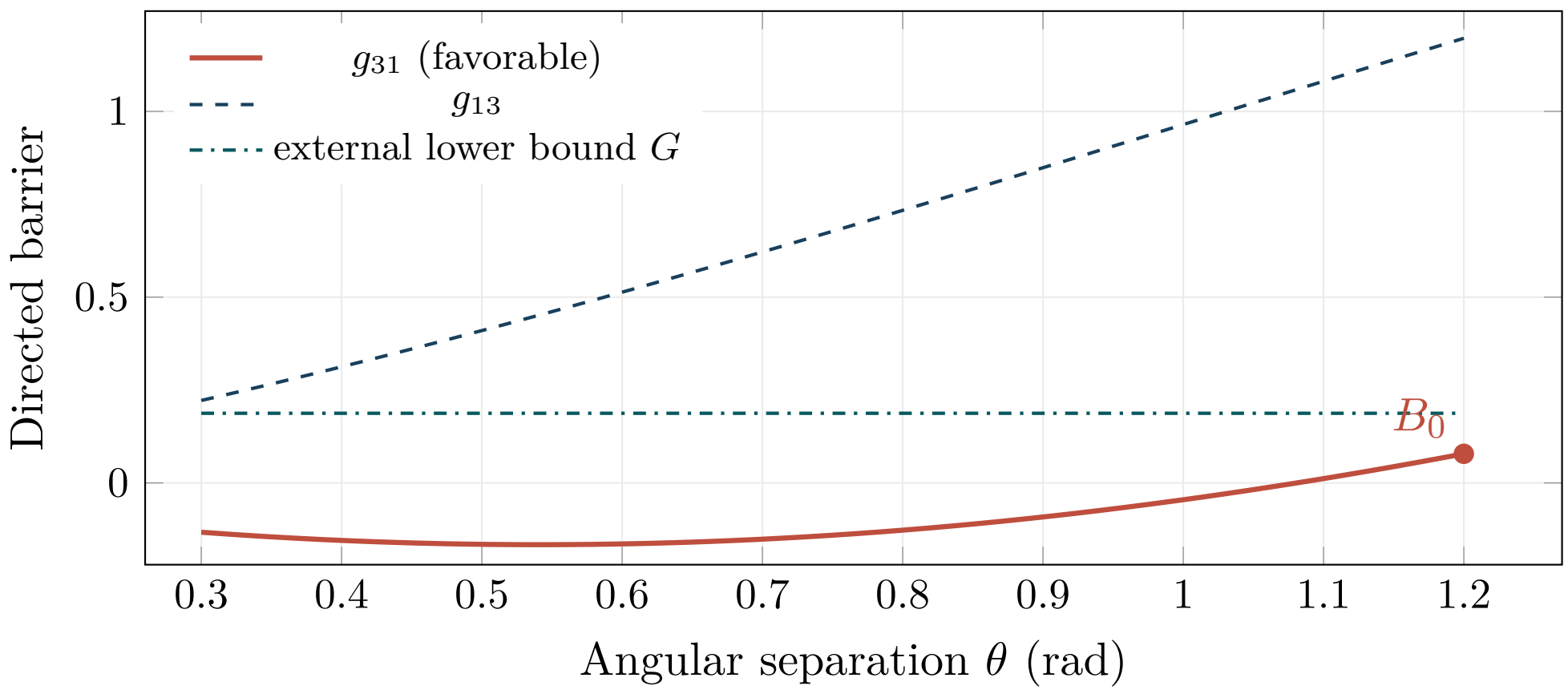
\begin{figure}[H]
\centering
\begin{tikzpicture}
\begin{axis}[width=.84\linewidth,height=6.15cm,xlabel={Angular separation $\theta$ (rad)},ylabel={Directed barrier}, xmin=.26,xmax=1.27,ymin=-.22,ymax=1.27,grid=both,grid style={draw=gray!15},legend style={font=\footnotesize,draw=none,fill=white,at={(.02,.98)},anchor=north west},tick label style={font=\small}]
\addplot[coral,very thick,domain=.3:1.2,samples=150]{1-cos(deg(x))-.6*sin(deg(x))};\addlegendentry{$g_{31}$ (favorable)}
\addplot[deep,thick,dashed,domain=.3:1.2,samples=150]{1-cos(deg(x))+.6*sin(deg(x))};\addlegendentry{$g_{13}$}
\addplot[teal!75!black,thick,dashdotted,domain=.3:1.2]{1-cos(deg(.8))*sqrt(1+.6^2)};\addlegendentry{external lower bound $G$}
\addplot[only marks,mark=*,mark size=2.2pt,coral] coordinates {(1.2,.0784187939)};
\node[above left,coral] at (axis cs:1.2,.0784187939) {$B_0$};
\end{axis}
\end{tikzpicture}
\caption{The \emph{entire corridor}, rather than the initial ordering alone, certifies the first event. The external barrier stays strictly above the favorable internal barrier's initial maximum.}
\label{fig:barriers}
\end{figure}

Along the isolated directed corridor with angular separation $\theta$, the favorable internal barrier is
$g_{31}(\theta)=1-\cos\theta-0.6\sin\theta$, and its maximum on $[0.3,1.2]$ is its value at $\theta=1.2$. Its opposite-directed counterpart is $g_{13}(\theta)=1-\cos\theta+0.6\sin\theta$.
Figure~\ref{fig:barriers} compares these two functions with the uniform
external-barrier lower bound in~\eqref{eq:Gexample}.

\begin{theorem}[Two separate exponential clocks in the explicit configuration]\label{thm:secondlower}
For the initial data~\eqref{eq:example} let $t_1(\beta)$ be the first crossing of angle $0.3$ and
\begin{equation}
t_2(\beta)=\inf\{t:\min(\theta_{12}(t),\theta_{23}(t))\le0.3\}.
\end{equation}
Then $t_2(\beta)<\infty$ for every $\beta>0$, and with $G=1-\cos(0.8)\sqrt{1+0.6^2}$,
\begin{equation}\label{eq:two-clocks}
\boxed{\liminf_{\beta\to\infty}\frac{\log t_2(\beta)}{\beta}\ge G,
\qquad \liminf_{\beta\to\infty}\frac{\log(t_2/t_1)}{\beta}\ge G-B_0>0.}
\end{equation}
Thus the first event and any event involving the external third point are separated on the exponential scale. This earlier lower bound is sharpened to the exact exponent by Corollary~\ref{cor:full-second} below.
\end{theorem}
\begin{proof}
The three initial scalar products are positive, so $t_2<\infty$ by \cref{thm:acute}. It remains to prove the lower bound. Let $\mathcal A$ denote the entire planar arc of angles $[0,1.2]$ in the $xy$-plane, and fix any $a<G$. In a small fixed neighborhood of configurations in which $u_1,u_3\in\mathcal A$ and $u_2$ retains its initial position, every external directed barrier is at least $G-\delta$ for arbitrarily small fixed $\delta>0$. Their total transition force is bounded by $\eta=C e^{-\beta(G-\delta)}$.

Here it is essential to control the two internal points for a time much longer than their first event. Write $z_i$ for their coordinates normal to the $xy$-plane, and let $E=\max(|z_1|,|z_3|)$. The internal portion of their equations is $\dot u_1=p_{13}(u_3-cu_1)$ and $\dot u_3=p_{31}(u_1-cu_3)$, where $c=u_1^\T u_3$. At any active maximum of $|z_i|$, positivity of $p_{ij}$ gives the upper Dini-derivative bound
\begin{equation}
D^+E\le q(1-c)E+C\eta,\quad q=p_{13}+p_{31}.
\end{equation}
Meanwhile $f=1-c$ satisfies $\dot f=-q(1-c^2)+R$, with $|R|\le C\eta$. As long as $c\ge0$,
\begin{equation}
\int_0^tq(s)(1-c(s))\,\dd s
\le\int_0^tq(s)(1-c(s)^2)\,\dd s
\le f(0)+C\eta t.
\end{equation}
Gronwall therefore yields $E(t)\le C\eta t\exp(f(0)+C\eta t)$.

The planar angular coordinates of $u_1,u_3$ satisfy cooperative equations: at the smaller angle the internal contribution is nonnegative, and at the larger it is nonpositive. The external contributions are $O(\eta)$ while the planar radii stay bounded away from zero. Thus the smaller angle remains above $-C\eta t$ and the larger below $1.2+C\eta t$. Also $\|u_2(t)-u_2(0)\|\le C\eta t$.  These estimates close a tubular bootstrap for $0\le t\le e^{\beta a}$ whenever $\delta<G-a$, since $\eta e^{\beta a}\to0$. Throughout this interval, $u_2$ stays at angular distance at least $0.8-o(1)>0.3$ from both other points. Consequently $t_2\ge e^{\beta a}$ for all large $\beta$. Letting $a\uparrow G$ proves the first inequality in~\eqref{eq:two-clocks}; the second follows from \cref{thm:corridor}.
\end{proof}

For the isolated internal pair, the boundary Laplace method yields the leading prefactor
\begin{equation}\label{eq:pref}
\boxed{\tau_{13}(0.3)\sim
\frac{e^{\beta B_0}}{\beta\sin(1.2)\,[\sin(1.2)-0.6\cos(1.2)]}
\approx\frac{1.50137}{\beta}e^{0.07841879\beta}.}
\end{equation}
The reason is that the maximal favorable barrier occurs uniquely at the initial endpoint and its derivative there is strictly positive. All external effects in the corridor are exponentially smaller, even after multiplication by $\beta$. A complete relative phase-speed comparison is needed to transfer this isolated endpoint calculation to the full ODE; the logarithmic first-event theorem alone does not imply the prefactor. Equation~\eqref{eq:pref} should therefore be read as an isolated-pair asymptotic and a full-flow prediction pending that comparison.

\FloatBarrier
\section{Rotating relative equilibria and the absence of universal coarsening}\label{sec:rotation}
Set $d=2$, $n=3$, equal masses, and $A=J$. At equally spaced angles
\begin{equation}
\theta_j(t)=\omega t+2\pi j/3,\qquad j=0,1,2,
\end{equation}
the angular ODE induced by~\eqref{eq:model} gives
\begin{equation}\label{eq:omega}
\boxed{\omega=-\frac{\sqrt3 e^{-\beta/2}
\sinh(\sqrt3\beta\eps/2)}{e^\beta+2e^{-\beta/2}\cosh(\sqrt3\beta\eps/2)}.}
\end{equation}
For $\eps\ne0$ this is a nonconstant periodic orbit, with $u_i^\T u_j=-1/2$ for all $i\ne j$. It is a rotating \emph{relative equilibrium}, not a merger. In particular no unconditional theorem claiming that all separated configurations coarsen can hold. Nor can the full vector field be the gradient of a time-independent scalar potential for a positive-definite Riemannian metric, because a nontrivial periodic orbit would contradict strict energy monotonicity.

\begin{figure}[H]
\centering
\begin{tikzpicture}[>=Latex,scale=1]
\draw[deep!65,thick] (0,0) circle (1.35);
\foreach \a/\j in {90/0,210/1,330/2}{
\fill[teal] (\a:1.35) circle (3pt);
\node[deep] at (\a:1.66) {$u_{\j}$};
\draw[coral,->,thick] (\a:1.7) arc[start angle=\a,end angle=\a-24,radius=1.7];
}
\draw[deep!60,dashed] (90:1.35)--(210:1.35)--(330:1.35)--cycle;
\node[deep,align=left] at (4.25,.65) {all pairwise angles\\remain $2\pi/3$};
\node[coral!90!black,align=left] at (4.25,-.64) {$\eps\ne0$: rigid rotation\\rather than collision};
\end{tikzpicture}
\caption{A nontrivial periodic orbit in the nonreciprocal three-body flow. Arrows indicate the direction for one sign of $\eps$; reversing $\eps$ reverses rotation.}
\label{fig:rotating}
\end{figure}
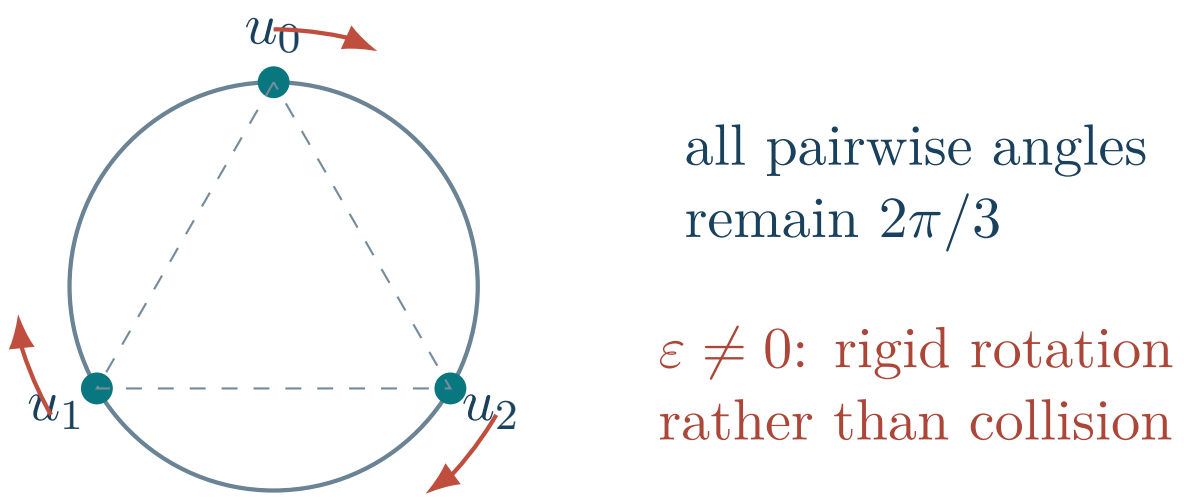

For $|\beta\eps|\ll1$, the angular velocity obeys
\begin{equation}
\omega=-\frac{3\beta\eps}{2}\frac{e^{-3\beta/2}}{1+2e^{-3\beta/2}}+O((\beta\eps)^3 e^{-3\beta/2}),
\end{equation}
exhibiting a potentially exponentially long \emph{rotation} time scale, distinct from an exponentially long merger time.
Figure~\ref{fig:rotating} depicts the resulting relative equilibrium.

\subsection{Exact rotating-triangle spectrum}
The angular equation is $\dot\theta_i=\sum_jp_{ij}\sin(\theta_j-\theta_i)$.
The rotating triangular relative equilibrium has an \emph{exact} spectrum;
its neutral mode must not be confused with asymptotic stability.
\begin{theorem}[Strict relative instability of the rotating triangle]\label{thm:rot-spectrum}
For every $\beta>0$ and $\eps\in\R$, the equilateral relative equilibrium on $\sphere^1$ has one zero angular eigenvalue (common rotation) and two eigenvalues with strictly positive real part.
Let
\begin{equation}
 r=\frac{\sqrt3}{2},\quad
 a=\frac{e^{-\beta(3/2+\eps r)}}{1+e^{-\beta(3/2+\eps r)}+e^{-\beta(3/2-\eps r)}},\quad
 b=\frac{e^{-\beta(3/2-\eps r)}}{1+e^{-\beta(3/2+\eps r)}+e^{-\beta(3/2-\eps r)}}.
\end{equation}
Then $\omega=r(a-b)$ and, with
\begin{align}
 \ell_+&=a\bigl[-\tfrac12+\beta(-r+\eps/2)(r-\omega)\bigr],\\
 \ell_-&=b\bigl[-\tfrac12+\beta(r+\eps/2)(-r-\omega)\bigr],
\end{align}
the eigenvalues are
\begin{equation}\label{eq:rot-spectrum}
 \boxed{\lambda_0=0,\qquad
 \lambda_{\pm}=-\tfrac32(\ell_++\ell_-)
 \pm i\tfrac{\sqrt3}{2}(\ell_+-\ell_-).}
\end{equation}
In fact $\operatorname{Re}\lambda_{\pm}\ge\tfrac34(a+b)>0$.
\end{theorem}
\begin{proof}
At each equilateral configuration, the two off-diagonal derivatives in the angular Jacobian are $\ell_+,\ell_-$. Rotational equivariance makes its diagonal entry $-(\ell_++\ell_-)$ and its circulant eigenvalues give~\eqref{eq:rot-spectrum}. A rearrangement yields
\begin{equation}
 \ell_++\ell_-=-\left(\tfrac12+\tfrac{3\beta}{4}\right)(a+b)
  +\beta\omega^2+\tfrac{\beta\eps\omega}{2}(1-a-b).
\end{equation}
Since $\eps\omega\le0$, $a+b\le1$, and $\omega^2=\tfrac34(a-b)^2\le\tfrac34(a+b)$, we conclude $\ell_++\ell_-\le-(a+b)/2$. The asserted strict real-part bound follows.
\end{proof}
\begin{remark}
On $\sphere^1$ with three points there are precisely three angular eigenvalues: one zero and the conjugate pair. There is no additional independent transverse mode. In the symmetric case $\eps=0$, the two nonzero eigenvalues coincide and are positive. Their smallness for large $\beta$ describes a \emph{weakly unstable}, rather than attracting, triangular orbit.
\end{remark}

\FloatBarrier
\section{What happens after the first event?}\label{sec:recursive}
The next question is whether the dynamics admits a rigorous recursive reduction. We can prove the \emph{exact} reduction on a collision manifold, but not yet the requisite shadowing of this manifold by near-collisions.

\begin{proposition}[Exact weighted quotient on collision strata]\label{prop:quotient}
Let $\cP=\{I_1,\ldots,I_k\}$ be a partition of $\{1,\ldots,n\}$. The stratum
\begin{equation}
\cM_{\cP}=\{U:u_i=v_a\ \text{for each }i\in I_a\}
\end{equation}
is flow-invariant. On this stratum the exact reduced equations for $v_a$ have the form~\eqref{eq:model}, with the \emph{combined} masses
\begin{equation}
\widetilde m_a=\sum_{i\in I_a}m_i.
\end{equation}
\end{proposition}
\begin{proof}
If $u_i=u_j$, the $i$th and $j$th rows of the attention kernel have the same denominator and induce the same velocity; uniqueness maintains equality.  Terms in the same block of a numerator add as
$\sum_{j\in I_b}m_j e^{\beta v_a^\T Mv_b}=\widetilde m_b e^{\beta v_a^\T Mv_b}$, giving the reduced system.
\end{proof}

\begin{proposition}[Uniform transverse linearization at complete consensus]\label{prop:linear}
At $U_*=(v,\ldots,v)$, with $\xi_i\in T_v\sphere^{d-1}$, the linearized flow is
\begin{equation}\label{eq:linear}
\dot\xi_i=\overline\xi-\xi_i,\qquad
\overline\xi=\frac{\sum_jm_j\xi_j}{\sum_jm_j}.
\end{equation}
The mean mode has eigenvalue zero and all $n-1$ transverse modes have eigenvalue $-1$, independently of $\beta,A,\eps$.
\end{proposition}
\begin{proof}
At consensus the weighted probabilities are $p_{ij}=m_j/\sum_km_k$. The derivative of $p_{ij}$ multiplies $v$, which is removed by $I-vv^\T$. The derivative of the projection contributes $-\xi_i$, while differentiation of the value vectors contributes $\bar\xi$. This gives~\eqref{eq:linear}.
\end{proof}

\FloatBarrier
\section{Partial-collision stability and a boundary-layer prefactor}\label{sec:refinement}
\subsection{What the normal exponent really is}
The eigenvalue $-1$ from \cref{prop:linear} is \emph{not} the exact normal eigenvalue of a partially collided block. Here is the correct general formula.
\begin{proposition}[Exact normal linearization of a two-point collision stratum]\label{prop:partial-normal}
At a configuration with $u_a=u_b=v$ (but other points arbitrary), let
\begin{equation}
 w_j=\frac{m_j e^{\beta v^\T M u_j}}{\sum_k m_k e^{\beta v^\T M u_k}},
 \qquad \mu=\sum_j w_j u_j,\qquad P_v=I-vv^\T.
\end{equation}
The covariant linearization of the relative perturbation $\xi\in T_v\sphere^{d-1}$ is
\begin{equation}\label{eq:normal-op}
 \boxed{\mathcal L_v\xi=-(v^\T\mu)\xi+
 \beta P_v \left[\sum_jw_j(u_j-\mu)(M(u_j-\mu))^\T\right]\xi.}
\end{equation}
If $\eta=\sum_{j\notin\{a,b\}}w_j$, then
\begin{equation}\label{eq:normal-bound}
 \boxed{\|\mathcal L_v+I\|\le2\eta+4\beta\|M\|_{\rm op}\eta.}
\end{equation}
Thus a shield with $(1+\beta)\eta=o(1)$ makes the normal eigenvalues $-1+o(1)$, not identically $-1$.
\end{proposition}
\begin{proof}
With the key configuration $U$ frozen, write
$f(x;U)=(I-xx^\T)\mu(x;U)$,
$\mu(x;U)=\sum_jw_j(x;U)u_j$. The difference of the two vector fields is exactly $f(u_a;U)-f(u_b;U)$, so key variations cancel to first order in the relative mode. Differentiating the row-wise softmax gives
\begin{equation}
 D_x\mu(v;U)=\beta\sum_jw_j(u_j-\mu)(M(u_j-\mu))^\T.
\end{equation}
The tangent projection of $D_x f(v;U)$ is~\eqref{eq:normal-op}. Since the internal key vectors are $v$, we have $1-v^\T\mu\le2\eta$ and
$\sum_jw_j\|u_j-\mu\|^2=1-\|\mu\|^2\le4\eta$. The operator bound follows.
\end{proof}
\begin{remark}[Correct Lyapunov scaling]
For $V=\tfrac12\|u_a-u_b\|^2$ and an exact normal law $\dot\xi=-\xi$, one has $\dot V=-2V$, not $-V$. More importantly, the exact collision stratum is invariant: there is no arbitrary additive external forcing in the \emph{relative} equation, because identical queries have identical velocities. A robust nonlinear tube bound still requires a uniform estimate of $D_xf$ in a neighborhood, as well as control of the quotient dynamics; \cref{prop:partial-normal} alone is not a global shadowing theorem.
\end{remark}

\subsection{The rotating inner pair produces a nonvanishing exponential bias}
Return to the three-point configuration~\eqref{eq:example} and \emph{first suppress its external point}. Write $u_1=(\cos\varphi,\sin\varphi,0)$ and $u_3=(\cos(\varphi+\theta),\sin(\varphi+\theta),0)$, starting at $(\varphi,\theta)=(0,1.2)$. The two internal transition probabilities are
\begin{equation}
 p_+(\theta)=\frac1{1+e^{\beta(1-\cos\theta+\eps\sin\theta)}},\qquad
 p_-(\theta)=\frac1{1+e^{\beta(1-\cos\theta-\eps\sin\theta)}}.
\end{equation}
Consequently $\dot\varphi=p_+\sin\theta$, $\dot\theta=-(p_++p_-)\sin\theta$. The final center angle of the isolated pair is exactly
\begin{equation}\label{eq:phi-integral}
 \Phi_\beta=\int_0^{1.2}\frac{p_+(\theta)}{p_+(\theta)+p_-(\theta)}\,\dd\theta.
\end{equation}
\begin{proposition}[A finite bias left by the inner boundary layer]\label{prop:phi}
For fixed $\eps>0$ and initial $\theta_0\in(0,\pi)$,
\begin{equation}\label{eq:phi-limit}
 \boxed{\lim_{\beta\to\infty}\beta\Phi_\beta=\frac{\log2}{\eps}.}
\end{equation}
\end{proposition}
\begin{proof}
Set $x=\beta\theta$ in~\eqref{eq:phi-integral}. For each fixed $x\ge0$, the numerator-to-total ratio converges to $(1+e^{\eps x})^{-1}$. The ratio is bounded by $2e^{-\beta\eps\sin\theta}$; since $\sin\theta\ge c_{\theta_0}\theta$ on $[0,\theta_0]$ for $\theta_0<\pi$, dominated convergence applies. Thus $\beta\Phi_\beta\to\int_0^\infty(1+e^{\eps x})^{-1}\dd x=(\log2)/\eps$.
\end{proof}
The displacement $\Phi_\beta=O(\beta^{-1})$ vanishes geometrically, but $\beta\eps\Phi_\beta\to\log2$ and hence its influence on a directed exponential weight does \emph{not} vanish.

\subsection{A rigorous prefactor for the exactly collapsed quotient}
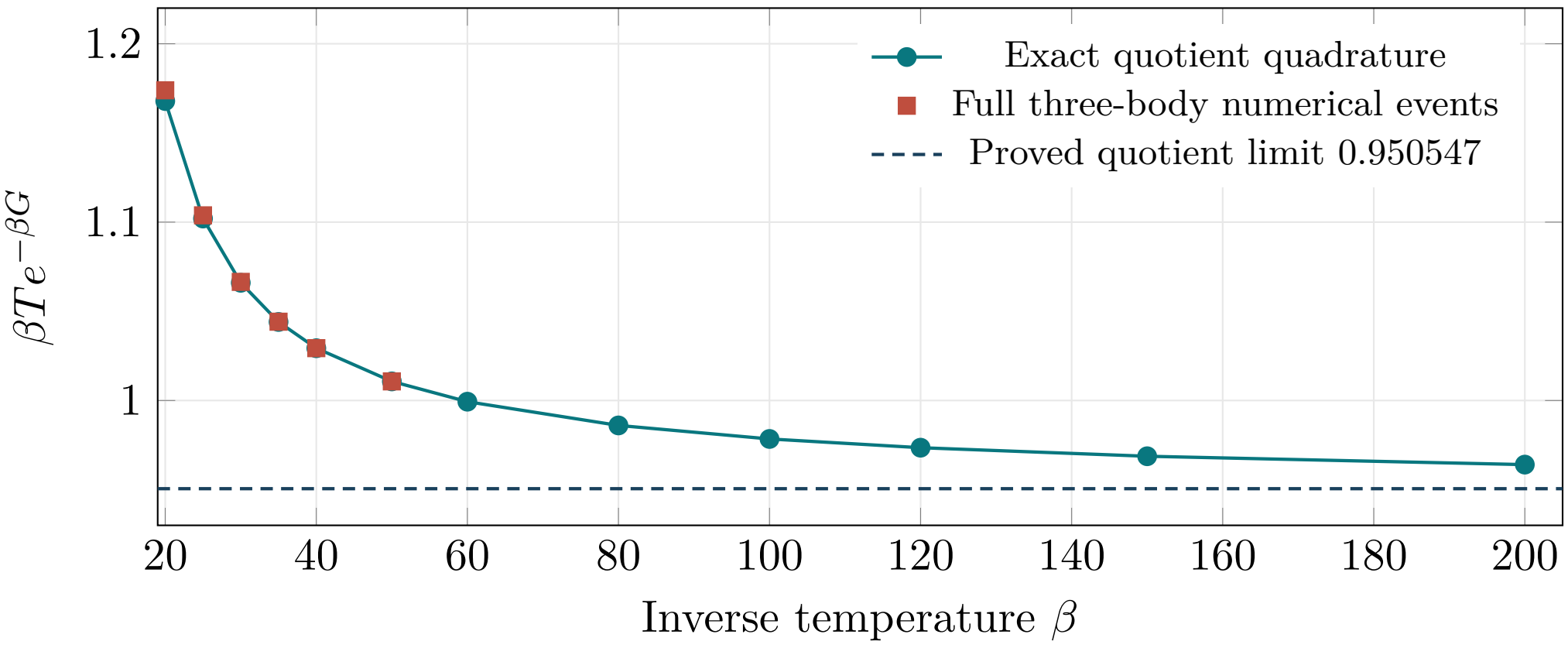
\begin{figure}[H]
\centering
\begin{tikzpicture}
\begin{axis}[width=.86\linewidth,height=6.0cm,xmin=19,xmax=205,ymin=.93,ymax=1.22,
 xlabel={Inverse temperature $\beta$},ylabel={$\beta T e^{-\beta G}$},grid=major,
 grid style={draw=gray!18},legend pos=north east,
 legend style={draw=none,fill=white,font=\footnotesize}]
\addplot[teal,mark=*,thick] coordinates {(20,1.167887)(25,1.102014)(30,1.065996)(35,1.044034)(40,1.029312)(50,1.010707)(60,.999330)(80,.986016)(100,.978436)(120,.973533)(150,.968740)(200,.964051)};
\addlegendentry{Exact quotient quadrature}
\addplot[coral,only marks,mark=square*,mark size=2pt] coordinates {(20,1.173906)(25,1.103809)(30,1.066502)(35,1.044172)(40,1.029349)(50,1.010709)};
\addlegendentry{Full three-body numerical events}
\addplot[deep,densely dashed,thick,domain=19:205] {0.95054656};
\addlegendentry{Proved quotient limit $0.950547$}
\end{axis}
\end{tikzpicture}
\caption{The boundary layer produces an order-one skew-weight correction. The numerically calculated exact quotient tends toward its proved limit, while finite-$\beta$ full dynamics closely follow the reduced clock. A proof of uniform full-flow matching remains open.}
\label{fig:quotient-prefactor}
\end{figure}

Let $v_\beta=(\cos\Phi_\beta,\sin\Phi_\beta,0)$ and $w=(\cos0.8,0,\sin0.8)$. Consider an \emph{exact} two-body quotient initialized at $(v_\beta,w)$, with masses $(2,1)$ and the same $M$. Let $T_\beta^{\rm red}$ denote its hitting time of angle $0.3$.
\begin{theorem}[Matched prefactor for the exact reduced model]\label{thm:matched-quotient}
For $\eps=0.6$ and the configuration above, set $c=\cos0.8$, $s=\sin0.8$, $G=1-c$. Then
\begin{equation}\label{eq:matched-pref}
 \boxed{\lim_{\beta\to\infty}\beta T_\beta^{\rm red}e^{-\beta G}
 =\frac{1}{s^2\left(\frac12 2^{c}+2\,2^{-c}\right)}
 \approx0.95054656.}
\end{equation}
\end{theorem}
\begin{proof}
The initial ordinary and skew scalar products satisfy
\begin{equation}
 c_\beta=c\cos\Phi_\beta=c+O(\beta^{-2}),\qquad
 h_\beta=c\sin\Phi_\beta=\frac{c\log2}{\eps\beta}+o(\beta^{-1}).
\end{equation}
Thus the two directed barriers at the initial endpoint are
$G\mp c(\log2)/\beta+o(\beta^{-1})$. With masses $(2,1)$, their two probabilities obey
\begin{equation}
 e^{\beta G}(q_{vw}+q_{wv})\longrightarrow
 K=\tfrac12 2^{c}+2\,2^{-c}.
\end{equation}
The two-body first integral~\eqref{eq:alpha} preserves the skew product to relative order $O(\beta^{-1})$ in the $O(\beta^{-1})$ endpoint layer. In~\eqref{eq:integral}, the maximal exponential barrier occurs at its initial endpoint with derivative $-1+O(\beta^{-1})$ in the $c$ coordinate. The endpoint Laplace method gives
$T_\beta^{\rm red}\sim e^{\beta G}/[\beta(1-c^2)K]$, proving the claim.
\end{proof}
\begin{remark}[Crucial scope distinction]
This theorem is \emph{exact for the constructed reduced two-body dynamics}. To identify $T_\beta^{\rm red}$ with the second-event clock $t_2$ of the original three-point flow up to $1+o(1)$, one must still prove quantitative post-event shadowing and a controlled event crossing. The observed diagnostic $Q_\beta=\beta t_2e^{-\beta G}$ approaching $1$ from above at the available $\beta$ does \emph{not} imply a limiting value of $1$: the reduced model predicts the different limit $0.95054656$.
\end{remark}
\begin{table}[H]
\centering\small
\caption{The full three-body numerical diagnostic compared to the exact two-body quotient quadrature; neither finite table implies equality of the asymptotic limits.}
\label{tab:quotient-diagnostic}
\begin{tabular}{r r r r}
\toprule
$\beta$ & $Q_\beta$ (full ODE) & $Q_\beta^{\rm red}$ (quadrature) & relative gap\\
\midrule
20&1.173906&1.167887&$5.15\times10^{-3}$\\
25&1.103809&1.102014&$1.63\times10^{-3}$\\
30&1.066502&1.065996&$4.75\times10^{-4}$\\
35&1.044172&1.044034&$1.32\times10^{-4}$\\
40&1.029349&1.029312&$3.60\times10^{-5}$\\
50&1.010709&1.010707&$3.0\times10^{-6}$\\
\bottomrule
\end{tabular}
\end{table}
Figure~\ref{fig:quotient-prefactor} extends the quotient quadrature to larger
$\beta$ and displays the distinction between the reduced-model limit and
finite full-flow measurements.

\subsection{The remaining nonlinear shadowing bridge}
The first threshold is defined at angle $0.3$, so at the event time itself
$V=1-\cos(0.3)$: the trajectory is \emph{not} yet within $o(\beta^{-1})$ of the collision stratum. A separate post-threshold relaxation interval is necessary. Once the directed inner rate is bounded below, reaching a polynomially small transverse scale takes $O(\log\beta)$ time; reaching an exponentially small one may take $O(\beta)$ time. Thus the fast time variable $\tau=\beta t$ cannot be asserted universally from the collision eigenvalue, which is of order one under a shield.

A promising route for proving equality between full and quotient prefactors is to use the \emph{integrated interaction rate} rather than a global Lipschitz estimate. For the exact two-body quotient, $\dot\psi=-q(\psi)\sin\psi$ gives
\begin{equation}
 \int_{t_0}^{T}q(\psi(t))\,\dd t
 =\int_{\psi(T)}^{\psi(t_0)}\frac{\dd\psi}{\sin\psi}=O(1)
\end{equation}
when $\psi$ stays away from $0,\pi$. Hence the integral of a Jacobian bound of order $\beta q$ is only $O(\beta)$ despite the exponential physical time interval. If a transverse discrepancy has already decreased to $e^{-K\beta}$ with sufficiently large $K$, the corresponding variation-of-constants estimate can potentially be closed. This is an \emph{approach}, not yet a verified full-flow shadowing theorem; the required coupling and event-transversality inequalities must be supplied.

\subsection{What the Gram pencil does not prove}
When $n=d$ and $U$ is invertible, $C^{-1}H=U^{-\T}AU^\T$ is similar to the \emph{fixed} matrix $A$. Thus its spectrum, for instance $\{i,-i,0\}$ when $A$ is the coordinate-plane rotation in three dimensions, is determined by $A$ itself and is not a new angular-momentum law or a merger-order selector. For $n<d$ the conserved pencil describes the restriction of $A$ to the time-independent row span; for $n>d$ it is singular and requires a reduced pencil. Finite-time congruence preserves rank, but a rank-deficient \emph{limit} is possible because the congruence transformations can become ill-conditioned as $t\to\infty$.

\begin{problem}[Rigorous second-merger shadowing]\label{prob:second}
Suppose a first pair has entered a neighborhood of its collision stratum, while the remaining directed barriers stay separated. Find verifiable geometric and spectral hypotheses that imply, on a time interval of length $\exp(\beta H)$,
\begin{equation}\label{eq:shadow}
\sup_{0\le t\le e^{\beta H}}
\operatorname{dist}(U(t),\cM_{\cP})=o(\beta^{-1}),
\end{equation}
and that the induced quotient trajectory is sufficiently accurate to preserve the identity and exponent of the \emph{second} threshold crossing. Can the hypotheses be propagated after a second crossing?
\end{problem}
The scale $o(\beta^{-1})$ is motivated by the exponential sensitivity $\exp(-\beta g)$ to barrier errors: an $O(\beta^{-1})$ displacement may induce an order-one multiplicative change in transition rates. Even if~\eqref{eq:shadow} holds, a separate \emph{transversality} condition is needed to ensure that a small perturbation of a trajectory preserves the identity of a near-simultaneous merger.

\subsection{Numerical two-event separation}
\begin{figure}[H]
\centering
\begin{tikzpicture}
\begin{semilogyaxis}[width=.86\linewidth,height=6.1cm,xlabel={Inverse temperature $\beta$},ylabel={Threshold hitting time (log scale)},xmin=10,xmax=52,ymin=1,ymax=200000,grid=both,grid style={draw=gray!15},legend pos=north west,legend style={draw=none,fill=white,font=\small}]
\addplot[teal,very thick,mark=*,mark size=2.2pt] coordinates {(12,2.0294)(16,1.9790)(20,1.9727)(25,2.0134)(30,2.1017)(35,2.2405)(40,2.4404)(50,3.1102)};
\addlegendentry{first event $t_1$}
\addplot[coral,very thick,mark=square*,mark size=2.2pt] coordinates {(12,4.6172)(16,10.2422)(20,25.2915)(25,86.6800)(30,317.9790)(35,1215.7801)(40,4778.0083)(50,77908.7874)};
\addlegendentry{second event $t_2$}
\end{semilogyaxis}
\end{tikzpicture}
\caption{Two distinct numerical clocks. The first clock is covered by \cref{thm:corridor}; the rapidly growing second clock still lacks a rigorous general geometric selection theorem.}
\label{fig:twoevents}
\end{figure}
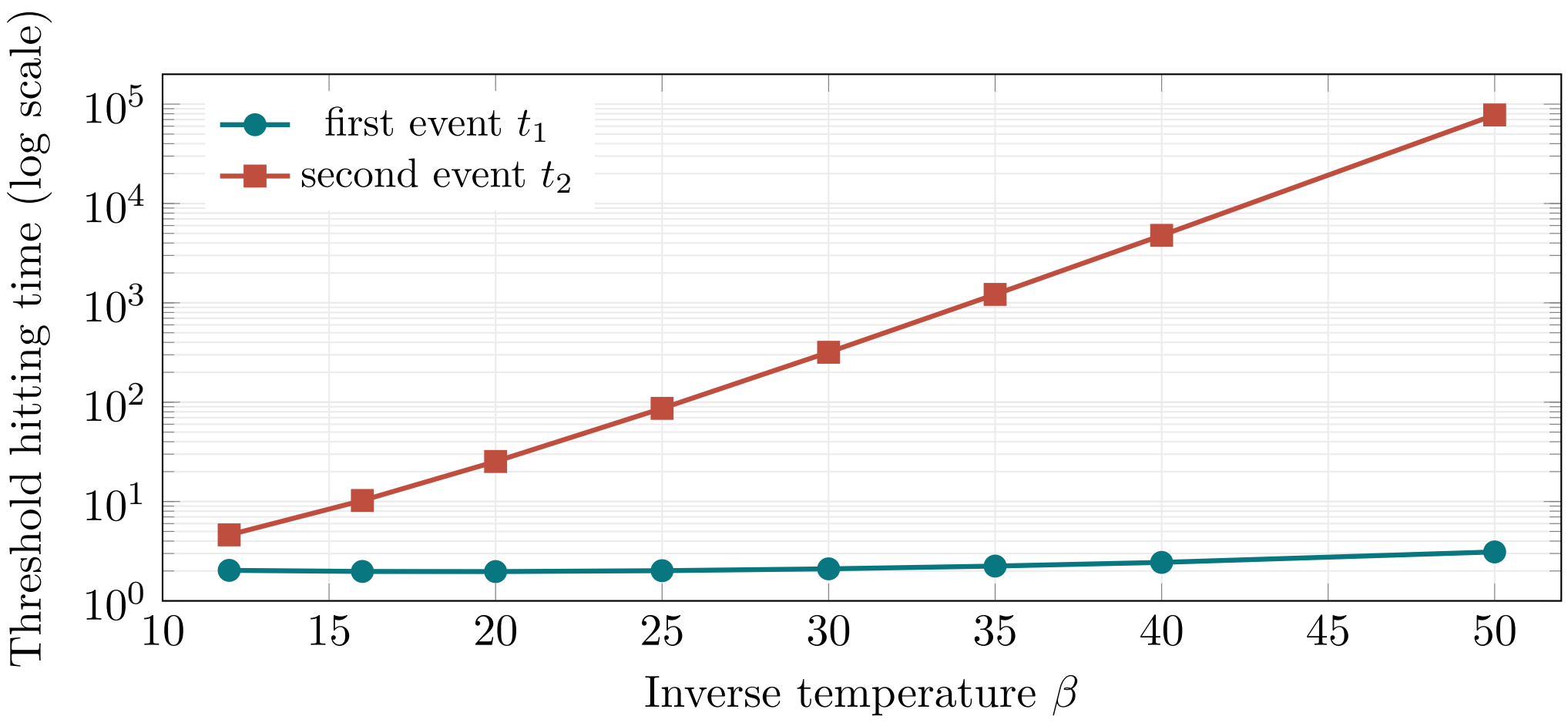

We integrated~\eqref{eq:model} in the concrete example~\eqref{eq:example} using a high-accuracy adaptive DOP853 solver with stable row-wise softmax and event detection at angle $0.3$. The \emph{first} event is $(1,3)$. The \emph{second} event is the first time another pair crosses the same threshold, with no forced collision or merging applied in the ODE.
Table~\ref{tab:events} and Figure~\ref{fig:twoevents} report the measured
separation of these two events.
\begin{table}[H]
\centering
\caption{Numerical threshold events for~\eqref{eq:example}; these data are not a theorem about the second-event asymptotic.}
\begin{tabular}{r r r r}
\toprule
$\beta$ & first event $t_1$ & second event $t_2$ & $t_2/t_1$\\
\midrule
12&2.0294&4.6172&2.28\\
16&1.9790&10.2422&5.18\\
20&1.9727&25.2915&12.82\\
25&2.0134&86.6800&43.05\\
30&2.1017&317.9790&151.29\\
35&2.2405&1215.7801&542.65\\
40&2.4404&4778.0083&1957.89\\
50&3.1102&77908.7874&25050.07\\
\bottomrule
\end{tabular}
\label{tab:events}
\end{table}

\subsection{The sharp second-clock exponent}
The exact weighted quotient in \cref{prop:quotient} suggests a specific exponent for the second event. If the first pair becomes a stable two-mass block at a point converging to $u_1(0)$ as $\beta\to\infty$, then its angle from the remaining point is asymptotically $0.8$, and its limiting directed skew product is zero. This motivated the formerly open identity, now proved in Corollary~\ref{cor:full-second},
\begin{equation}\label{eq:secondconj}
\boxed{\lim_{\beta\to\infty}\frac{\log t_2(\beta)}{\beta}
=1-\cos0.8\approx0.30329329.}
\end{equation}
The diagnostic $Q_\beta=\beta t_2\exp[-\beta(1-\cos0.8)]$ takes the values $1.174$, $1.067$, $1.029$ and $1.011$ at $\beta=20,30,40,50$. The rigorously calculated \emph{exact-quotient} prefactor in \cref{thm:matched-quotient} is $0.95054656$, not $1$ and not the naive zero-skew quotient value $1/[\frac52\sin^2(0.8)]\approx0.77730$. That numerical match, even at very high relative accuracy, does not yet prove that the full three-body $t_2$ has the same prefactor. The full-flow second exponent is now given by Corollary~\ref{cor:full-second}; the prefactor remains unresolved.

\FloatBarrier
\section{A proof audit of second-clock preservation}\label{sec:clock-audit}
The claim that a small normal distance from a collision stratum automatically preserves the next hitting-time prefactor is false. We separate a \emph{proved transverse estimate}, a \emph{proved conditional scalar comparison}, and the still unresolved tangential coupling of the full and quotient dynamics. This distinction is essential when the desired relative error is $1+o(1)$ rather than merely the correct exponential rate.

\subsection{A nonlinear contraction bound in a shielded tube}
Fix an ordered pair of indices $a\ne b$, put $u=u_a$, $v=u_b$, and let
\begin{equation}\label{eq:new-V}
V(t)=1-u(t)^\T v(t)=\frac12\|u(t)-v(t)\|^2,
\qquad K=\|M\|_{\mathrm{op}}\ge 1.
\end{equation}
For a frozen configuration $U$, extend the attention law to query points $x\in\R^d$ by
\begin{equation}\label{eq:query-map}
w_j(x;U)=\frac{m_j e^{\beta x^\T M u_j}}{\sum_{\ell}m_\ell e^{\beta x^\T M u_\ell}},\qquad
\mu(x;U)=\sum_j w_j(x;U)u_j.
\end{equation}
Define the \emph{segment shield}
\begin{equation}\label{eq:segment-shield}
\eta(t)=\sup_{0\le s\le1}\sum_{j\notin\{a,b\}}
w_j(su(t)+(1-s)v(t);U(t)).
\end{equation}
Note that this is an assumption on all intermediate queries, stronger than checking the external weights only at $u$ and $v$.

\begin{theorem}[Explicit nonlinear transverse contraction under a segment shield]\label{thm:nonlinear-tube}
Suppose $\beta\ge1$ and, for an interval $[t_0,T]$, the exact solution of~\eqref{eq:model} obeys
\begin{equation}\label{eq:tube-hyp}
V(t_0)\le\frac{1}{32\beta K},\qquad
\eta(t)\le\frac{1}{64(1+\beta K)}\quad(t\in[t_0,T]).
\end{equation}
Then
\begin{equation}\label{eq:tube-contraction}
\boxed{\;V(t)\le V(t_0)e^{-(t-t_0)},\qquad
\|u(t)-v(t)\|\le\sqrt{2V(t_0)}e^{-(t-t_0)/2}\;}
\end{equation}
for every $t\in[t_0,T]$. The time $T$ may depend on $\beta$ and may be exponentially large; the only long-time assumption is that the segment shield continues to hold.
\end{theorem}
\begin{proof}
Let $D=u-v$, and suppress the fixed configuration $U(t)$ in the notation. Direct differentiation of~\eqref{eq:new-V} using~\eqref{eq:model} gives the \emph{exact} identity
\begin{equation}\label{eq:exact-V}
\dot V=D^\T[\mu(u)-\mu(v)]
 -V\,[u^\T\mu(u)+v^\T\mu(v)].
\end{equation}
For a query $x$, the softmax Jacobian is the covariance operator
\begin{equation}\label{eq:covariance}
D_x\mu(x;U)=\beta\biggl(\sum_jw_j(x)u_ju_j^\T-\mu(x)\mu(x)^\T\biggr)M^\T.
\end{equation}
On the segment between $u$ and $v$, use $v$ as a comparison center. Since the covariance is positive semidefinite and its operator norm does not exceed its trace,
\begin{align}
\left\|\sum_jw_j(x)u_ju_j^\T-\mu(x)\mu(x)^\T\right\|_{\mathrm{op}}
&\le\sum_jw_j(x)\|u_j-v\|^2\notag\\
&\le 2V+4\eta(t).\label{eq:cov-bound}
\end{align}
By the mean-value formula,
\begin{equation}\label{eq:mu-diff}
\|\mu(u)-\mu(v)\|\le\beta K(2V+4\eta)\|D\|.
\end{equation}
The total weight on the two internal keys is at least $1-\eta$, each internal key has inner product at least $1-V$ with its own endpoint query, and external keys have inner product at least $-1$. Hence
\begin{equation}\label{eq:self-align}
u^\T\mu(u)\ge1-V-2\eta,\qquad
v^\T\mu(v)\ge1-V-2\eta.
\end{equation}
Combining~\eqref{eq:exact-V}--\eqref{eq:self-align} and $\|D\|^2=2V$ yields
\begin{equation}\label{eq:V-differential}
\dot V\le\left[-2+2V+4\eta+4\beta K V+8\beta K\eta\right]V.
\end{equation}
As long as $V\le 1/(32\beta K)$, the hypotheses~\eqref{eq:tube-hyp} make the bracket strictly smaller than $-1$. The resulting decay preserves the assumed tube bound, closing the first-exit bootstrap. Integrating $\dot V\le-V$ proves~\eqref{eq:tube-contraction}.
\end{proof}
\begin{remark}[What this theorem does and does not establish]\label{rem:tube-scope}
The exact collision stratum is invariant, so the relative equation has no compulsory additive $O(\eta)$ forcing term. Theorem~\ref{thm:nonlinear-tube} is a genuine uniform-in-time \emph{normal} estimate. It does not establish the segment shield for the target three-body trajectory, nor does it compare the motion \emph{along} the collision stratum with the quotient trajectory. The initial threshold $0.3$ in our example has $V=1-\cos(0.3)\approx0.04466$, which for arbitrarily large $\beta$ does not satisfy~\eqref{eq:tube-hyp}; a separate entrance/relaxation estimate is required. Once inside the tube, an additional $3\log\beta$ of contraction gives $\|u-v\|=O(\beta^{-2})$; obtaining $e^{-c\beta}$ from a polynomially small initial distance generally requires an $O(\beta)$ interval, not merely $O(\log\beta)$.
\end{remark}

\subsection{Normal proximity alone cannot determine a prefactor}
The next proposition gives a \emph{counterexample} to inferring second-clock preservation from $\operatorname{dist}(U,\cM_{\cP})=o(\beta^{-1})$ alone, even when that distance is exactly zero.
\begin{proposition}[Tangential $\beta^{-1}$ shifts change a metastable clock]\label{prop:prefactor-counterexample}
Consider the symmetric two-center quotient ($\eps=0$) with fixed positive masses $m_1,m_2$, initial angular separation $\psi_0\in(0,\pi)$, and fixed threshold $\psi_*\in(0,\psi_0)$. Let $T_\beta(\psi_0)$ be its hitting time, and let $a\in\R$ be fixed. Then
\begin{equation}\label{eq:shift-clock}
\boxed{\lim_{\beta\to\infty}\frac{T_\beta(\psi_0+a/\beta)}{T_\beta(\psi_0)}
= e^{a\sin\psi_0}.}
\end{equation}
Both initial states may be realized \emph{exactly} on the same two-block collision stratum. Their normal distances to that stratum vanish identically.
\end{proposition}
\begin{proof}
The angular equation of the symmetric quotient is
$\dot\psi=-q_\beta(\psi)\sin\psi$, with
\begin{equation}
q_\beta(\psi)=\left(\frac{m_2}{m_1}+\frac{m_1}{m_2}\right)e^{-\beta(1-\cos\psi)}(1+o(1))
\end{equation}
uniformly in a fixed neighborhood of $\psi_0$. Endpoint Laplace asymptotics therefore give, with $k=m_2/m_1+m_1/m_2$,
\begin{equation}\label{eq:sym-clock}
T_\beta(\psi_0)\sim\frac{e^{\beta(1-\cos\psi_0)}}{\beta k\sin^2\psi_0}.
\end{equation}
Since
$\beta[\cos\psi_0-\cos(\psi_0+a/\beta)]\to a\sin\psi_0$,
taking the ratio proves~\eqref{eq:shift-clock}. Exact block collisions supply quotient initial conditions lying on the stratum with zero transverse distance.
\end{proof}

\begin{figure}[H]
\centering
\begin{tikzpicture}[>=Latex,scale=1]
\fill[soft!65] (-.15,-1.18) rectangle (10.35,1.85);
\fill[teal!12] (-.15,-.12) rectangle (10.35,.12);
\draw[deep,very thick,->] (-.05,0)--(10.4,0) node[right]{$z$: quotient phase};
\draw[deep,very thick,->] (.2,-1.08)--(.2,1.46) node[above]{$r$: normal displacement};
\node[teal!70!black,above] at (8.05,.12) {$\cM_{\cP}:\ r=0$};
\draw[coral,very thick,-{Latex}] (2.2,1.18) .. controls (2.35,.65) and (2.6,.3) .. (3.03,.055);
\node[coral!90!black,anchor=west,scale=.9] at (2.4,1.07) {normal contraction};
\fill[deep] (5,0) circle (2.5pt) node[below=7pt] {$z_0$};
\fill[coral] (7.05,0) circle (2.5pt) node[below=7pt] {$z_0+a/\beta$};
\draw[coral,thick,{Latex}-{Latex}] (5.12,.45)--(6.92,.45)
  node[midway,above=3pt] (phase-shift-label) {$a/\beta$};
\node[anchor=east,align=right,deep,font=\small]
  (normal-distance-label) at (10.1,1.26) {same normal distance:\\zero};
\end{tikzpicture}
\caption{Normal control and tangential clock control are different. Two configurations can lie exactly on the collision stratum yet have different leading hitting-time constants because their quotient phases differ by $O(\beta^{-1})$. Schematic, not a numerical trajectory.}
\label{fig:tangential-obstruction}
\end{figure}
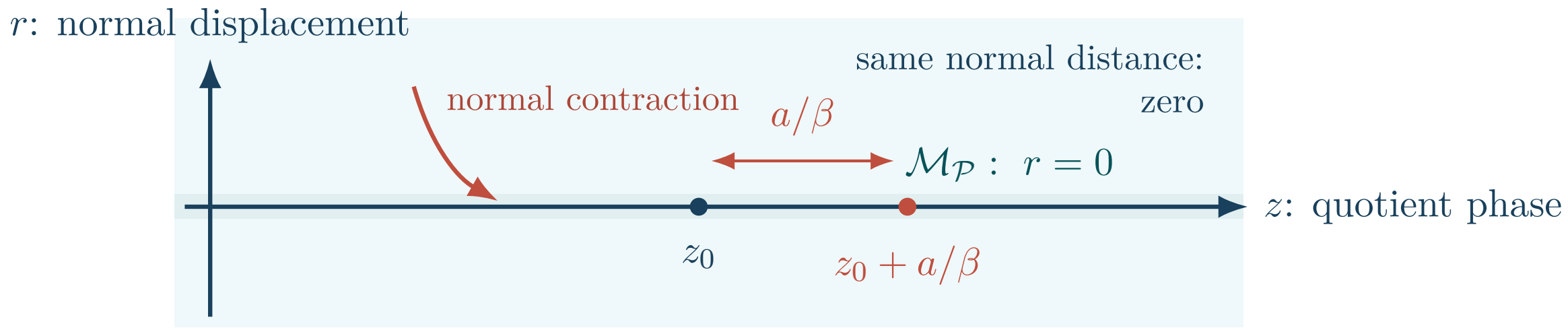

Figure~\ref{fig:tangential-obstruction} illustrates the two initial phases
in Proposition~\ref{prop:prefactor-counterexample}, both on the exact collision
stratum.

\subsection{A correct conditional criterion for relative clock preservation}
It is useful to formulate the scalar step that a future full-flow shadowing theorem would actually need to verify. Unlike the desired full-to-quotient reduction, the following is elementary and rigorous.
\begin{lemma}[Phase-velocity comparison and endpoint stability]\label{lem:phase-velocity}
Fix $0<\psi_*<\psi_0<\pi$. Suppose two \emph{monotonically decreasing} angular phases traverse $[\psi_*,\psi_0]$ with autonomous positive speed functions $F_\beta(\psi)$ and $\bar F_\beta(\psi)$, and
\begin{equation}\label{eq:velocity-relative}
\rho_\beta:=\sup_{\psi\in[\psi_*,\psi_0]}
\left|\frac{F_\beta(\psi)}{\bar F_\beta(\psi)}-1\right|\longrightarrow0.
\end{equation}
If both start at $\psi_0$, their hitting times satisfy $T_\beta/\bar T_\beta=1+O(\rho_\beta)$. The same conclusion holds for an initial phase perturbation $o(\beta^{-1})$, provided, on a fixed endpoint neighborhood, $\bar F_\beta(\psi)=a_\beta(\psi)e^{-\beta f(\psi)}$, where $a_\beta$ is uniformly bounded above and below by positive constants on the entire phase interval and locally regular at the upper endpoint, $f\in C^2$ has its unique global maximum on the interval at the upper endpoint, and $f'(\psi_0)>0$.
\end{lemma}
\begin{proof}
With common endpoints, $T_\beta=\int_{\psi_*}^{\psi_0}F_\beta(\psi)^{-1}\,\dd\psi$ and $\bar T_\beta=\int_{\psi_*}^{\psi_0}\bar F_\beta(\psi)^{-1}\,\dd\psi$. By~\eqref{eq:velocity-relative},
$\bar T_\beta/(1+\rho_\beta)\le T_\beta\le\bar T_\beta/(1-\rho_\beta)$ for large $\beta$. With an upper endpoint shift $\delta_\beta=o(\beta^{-1})$, the additional endpoint interval contributes at most $O(|\delta_\beta|e^{\beta f(\psi_0)+O(\beta|\delta_\beta|)})$, whereas endpoint Laplace gives $\bar T_\beta\asymp\beta^{-1}e^{\beta f(\psi_0)}$. The relative difference is $O(\beta|\delta_\beta|)=o(1)$.
\end{proof}
\begin{remark}[A phase-coordinate theorem, not an assumption-free full-flow result]
The original full three-body system is not an autonomous scalar angular equation. To invoke Lemma~\ref{lem:phase-velocity}, one must first control the tangential quotient coordinates, guarantee monotonicity, and show that the full phase speed \emph{at a given phase} is close to the quotient speed. Knowing only that $\|u_a-u_b\|=o(\beta^{-1})$ does not establish~\eqref{eq:velocity-relative}. Even if a relative speed error is merely $o(1)$, the lemma yields $1+o(1)$ timing accuracy, not the stronger $o(\beta^{-1})$ relative timing error claimed in a proposed proof.
\end{remark}

\subsection{The remaining quantitative tangential estimate}\label{subsec:tangential-gap}
Let $r(t)$ denote the normal error and $z(t)$ a properly synchronized tangential separation of the projected full trajectory and the quotient. A natural but \emph{not yet verified} differential inequality would be
\begin{equation}\label{eq:tangential-target}
D^+r\le-\kappa r,\qquad D^+z\le L_\beta(t)z+K_\beta(t)r.
\end{equation}
If it were established along the entire phase interval up to the second event, the standard variation-of-constants estimate would give
\begin{equation}\label{eq:tangential-criterion}
z(t)\le\exp\!\left(\int_{t_0}^{t}L_\beta^+(s)\,\dd s\right)
\left[z(t_0)+\int_{t_0}^{t}K_\beta(s)r(s)\,\dd s\right].
\end{equation}
A sufficient \emph{additional} requirement for $o(\beta^{-1})$ quotient closeness is that the right-hand side be $o(\beta^{-1})$ uniformly up to the second event. In an exactly monotone two-body quotient, $\int q\,\dd t=\int \dd\psi/\sin\psi=O(1)$ on an interval bounded away from $0$ and $\pi$, but $\int \beta q\,\dd t=O(\beta)$, not $O(1)$. This observation alone does not verify~\eqref{eq:tangential-target} for the full system or guarantee a sufficiently small prefactor in~\eqref{eq:tangential-criterion}.

One must also distinguish two different barriers in the concrete example. The initial second-clock exponent candidate is $G_*=1-\cos(0.8)\approx0.30329329$, whereas the earlier \emph{uniform external shield lower bound} in the first corridor is $G_{\rm sh}\approx0.187507$. Neither can be used unchanged as a uniform bound on the entire subsequent trajectory: as the second angle approaches $0.3$, its symmetric base barrier becomes $1-\cos(0.3)\approx0.044664$. In addition, a tube estimate only through $T=e^{\beta H}$ for a fixed $H<G_*$ ends exponentially earlier than a clock of size $\beta^{-1}e^{\beta G_*}$. The endpoint derivative $f'(\psi_0)>0$ controls Laplace asymptotics but does not supply a $\beta$-uniform lower bound on the physical crossing speed, which itself can be exponentially small.

\begin{problem}[Full-flow tangential matching and second-event prefactor]\label{prob:full-clock}
For the three-point configuration~\eqref{eq:example}, prove the coupled bounds needed to apply Lemma~\ref{lem:phase-velocity} up to the second threshold and determine whether
\begin{equation}\label{eq:full-target}
\boxed{\lim_{\beta\to\infty}\beta t_2(\beta)e^{-\beta(1-\cos0.8)}
\stackrel{?}{=}\frac{1}{\sin^2(0.8)\left(\tfrac12 2^{\cos0.8}+2\,2^{-\cos0.8}\right)}.}
\end{equation}
The right-hand side is \emph{proved for the exactly collapsed two-body quotient only} by Theorem~\ref{thm:matched-quotient}. The full-flow identity remains unproved. A proposed recursive merger theorem must additionally propagate its geometric gap and synchronization assumptions across each stage.
\end{problem}

\FloatBarrier
\section{Collapse-layer estimates and a conditional phase-matching mechanism}\label{sec:collapse-audit}
We refine the proof checkpoints from \cref{sec:clock-audit}.  Two different claims must be kept apart: estimates for a perturbed \emph{two-center normal form}, which can be proved under explicit error bounds, and verification that the original three-dimensional flow satisfies those error bounds with a sufficiently accurate moving coordinate chart. Only the former is completed here.

\subsection{An almost-conserved skew invariant with an explicit stopping scale}
Let $u,v\in\sphere^{d-1}$ and write $c=u^\T v$, $h=u^\T Av$, $s=\sqrt{1-c^2}$ and $\alpha=h/s$ whenever $s>0$. Consider the perturbed two-center equations
\begin{equation}\label{eq:perturbed-pair}
\dot c=q(1-c^2)+r_c,\qquad \dot h=-qch+r_h,
\qquad |r_c|+|r_h|\le C\eta_\beta(t).
\end{equation}
The coefficient $q$ is nonnegative and may depend on the full configuration.  The exact two-center cancellation survives in the following sense.
\begin{lemma}[Skew invariant under external forcing]\label{lem:alpha-drift}
Assume \eqref{eq:perturbed-pair} with $|\alpha|\le\|A\|_{\rm op}$, as is automatically true for points on the sphere. Then
\begin{equation}\label{eq:alpha-drift}
\dot\alpha=\frac{r_h}{s}+\frac{ch\,r_c}{s^3},\qquad
|\dot\alpha|\le C\eta_\beta(t)
\left(\frac1s+\frac{\|A\|_{\rm op}}{s^2}\right).
\end{equation}
In particular, if $s(t)\ge\beta^{-N}$ for $t\in[t_a,t_b]$ and
$\eta_\beta(t)\le C_0 e^{-\beta G_{\rm sh}}$, then
\begin{equation}\label{eq:alpha-stopping}
|\alpha(t_b)-\alpha(t_a)|
\le C_1(t_b-t_a)(\beta^N+\beta^{2N})e^{-\beta G_{\rm sh}}.
\end{equation}
Thus an $O(\log\beta)$ collapse window stopped at $s=\beta^{-3}$ has exponentially small skew-invariant drift, conditional on the stated external shield.
\end{lemma}
\begin{proof}
Differentiate $h/s$ and use $\dot s=-c\dot c/s$. The terms involving $q$ cancel identically. Since the restriction of a skew form to the plane spanned by $u$ and $v$ satisfies $|u^\T Av|\le \|A\|_{\rm op}\sqrt{1-c^2}$, one has $|\alpha|\le \|A\|_{\rm op}$. Substituting $|h|\le\|A\|_{\rm op}s$ into the remaining terms yields \eqref{eq:alpha-drift}; integration gives \eqref{eq:alpha-stopping}.
\end{proof}
\begin{remark}[Scope of the $O(\log\beta)$ collapse estimate]
For the isolated pair of \eqref{eq:example}, the favorable barrier
$f(\theta)=1-\cos\theta-0.6\sin\theta$ is strictly negative on $(0,0.3]$. Hence the favorable transition probability is at least $1/2$, and the isolated $q$ is at least $1/2$ on that interval. A \emph{perturbed} solution inherits a positive rate only if its external interactions and its angular drift are controlled. Under such a bootstrap one obtains a logarithmic relaxation from the $0.3$ threshold to a polynomially small angular scale. This does not itself prove that the requisite shield holds for the full system throughout that window.
\end{remark}

\subsection{Angular reparameterization removes the physical metastable clock}
The next lemma is an elementary but useful mechanism for relative phase matching. It deliberately specifies all the assumptions needed for its scalar application.
\begin{lemma}[Conditional phase-defect estimate]\label{lem:angular-phase}
Let $0<\theta_{\min}<\theta_0<\pi$ and $p,q$ be nonnegative functions on $[\theta_{\min},\theta_0]$ with $p\le q$ and $q>0$. Suppose a perturbed planar pair obeys
\begin{equation}\label{eq:phase-perturbed}
\dot\theta=-q(\theta)\sin\theta+r_\theta(t),\qquad
\dot\varphi=p(\theta)\sin\theta+r_\varphi(t)
\end{equation}
while $\theta$ decreases strictly from $\theta_0$ to $\theta_{\min}$, and
$|r_\theta|+|r_\varphi|\le\eta(\theta)$ with
$\eta(\theta)\le\tfrac12 q(\theta)\sin\theta$.
Let $\varphi_{\rm iso}$ solve the unperturbed angular system with the same initial $\varphi$ and the same initial $\theta$. Then at any common angular coordinate,
\begin{equation}\label{eq:phase-defect}
\left|\frac{\dd\varphi}{\dd\theta}+\frac{p(\theta)}{q(\theta)}\right|
\le\frac{4\eta(\theta)}{q(\theta)\sin\theta},\qquad
|\varphi-\varphi_{\rm iso}|
\le4\int_{\theta}^{\theta_0}
\frac{\eta(a)}{q(a)\sin a}\,\dd a.
\end{equation}
\end{lemma}
\begin{proof}
By \eqref{eq:phase-perturbed},
$\dd\varphi/\dd\theta=-(p\sin\theta+r_\varphi)/(q\sin\theta-r_\theta)$. Subtract $-p/q$ and use $p\le q$ and $|r_\theta|\le\frac12q\sin\theta$. The numerator is bounded by $q(|r_\theta|+|r_\varphi|)$ and the denominator below by $\frac12q^2\sin\theta$, yielding the displayed factor $4$ as a valid conservative constant. Integrate in the monotone angular coordinate.
\end{proof}
\begin{corollary}[Phase bias under two separated error regions]\label{cor:phase-regions}
Under the hypotheses of \cref{lem:angular-phase}, let $\theta_{\min}=\beta^{-3}$ and $\theta_0=1.2$. Suppose, with constants independent of $\beta$, that
\begin{equation}\label{eq:regional-ratios}
\frac{\eta(\theta)}{q(\theta)}\le
\begin{cases}
C e^{-\beta\Delta},&0.3\le\theta\le1.2,\\
C e^{-\beta G_{\rm sh}},&\beta^{-3}\le\theta\le0.3,
\end{cases}
\qquad \Delta,G_{\rm sh}>0.
\end{equation}
Then the discrepancy from the \emph{complete} isolated terminal phase $\Phi_\beta$ obeys
\begin{equation}\label{eq:phase-terminal}
|\varphi(\beta^{-3})-\Phi_\beta|
\le C' e^{-\beta\Delta}+C'e^{-\beta G_{\rm sh}}\log\beta+\beta^{-3}=o(\beta^{-1}).
\end{equation}
Here the common initial phase is set to zero; the sign convention for $\varphi$ is that of \eqref{eq:phi-integral}.
\end{corollary}
\begin{proof}
On the fixed outer interval, $\int_{0.3}^{1.2}(\sin\theta)^{-1}\dd\theta$ is finite. On $[\beta^{-3},0.3]$, $\int(\sin\theta)^{-1}\dd\theta=O(\log\beta)$. Apply \eqref{eq:phase-defect} and \eqref{eq:regional-ratios}. The residual isolated phase from $0$ to $\beta^{-3}$ is at most $\beta^{-3}$ because $0\le p/q\le1$.
\end{proof}

\begin{figure}[H]
\centering
\begin{tikzpicture}
\begin{axis}[width=.88\linewidth,height=5.7cm,xmin=0,xmax=1.23,ymin=-.3,ymax=.34,
 xlabel={Internal angle $\theta$ (radians)},ylabel={Directed barrier},
 grid=both,grid style={draw=gray!13},legend style={at={(.02,.98)},anchor=north west,draw=none,fill=white,font=\footnotesize},samples=180]
\addplot[fill=teal!10,draw=none] coordinates {(.003,-.3)(.3,-.3)(.3,.34)(.003,.34)}\closedcycle;
\addplot[coral,thick,domain=0:1.2] {1-cos(deg(x))-0.6*sin(deg(x))};
\addlegendentry{favorable inner barrier $f(\theta)$}
\addplot[deep,densely dashed,thick,domain=0:1.2] {.18750734};
\addlegendentry{first-corridor shield $G_{\rm sh}$}
\draw[teal,thick,densely dashed] (axis cs:.3,-.3)--(axis cs:.3,.34);
\node[teal!70!black,anchor=south west,font=\footnotesize] at (axis cs:.315,-.26) {threshold $0.3$};
\end{axis}
\end{tikzpicture}
\caption{The favorable isolated inner barrier becomes negative in the post-threshold interval (shaded). Its rate is no longer exponentially slow there. The horizontal shield is a \emph{geometric bound from the initial corridor}, not a proved uniform shield until the second event.}
\label{fig:collapse-barrier}
\end{figure}
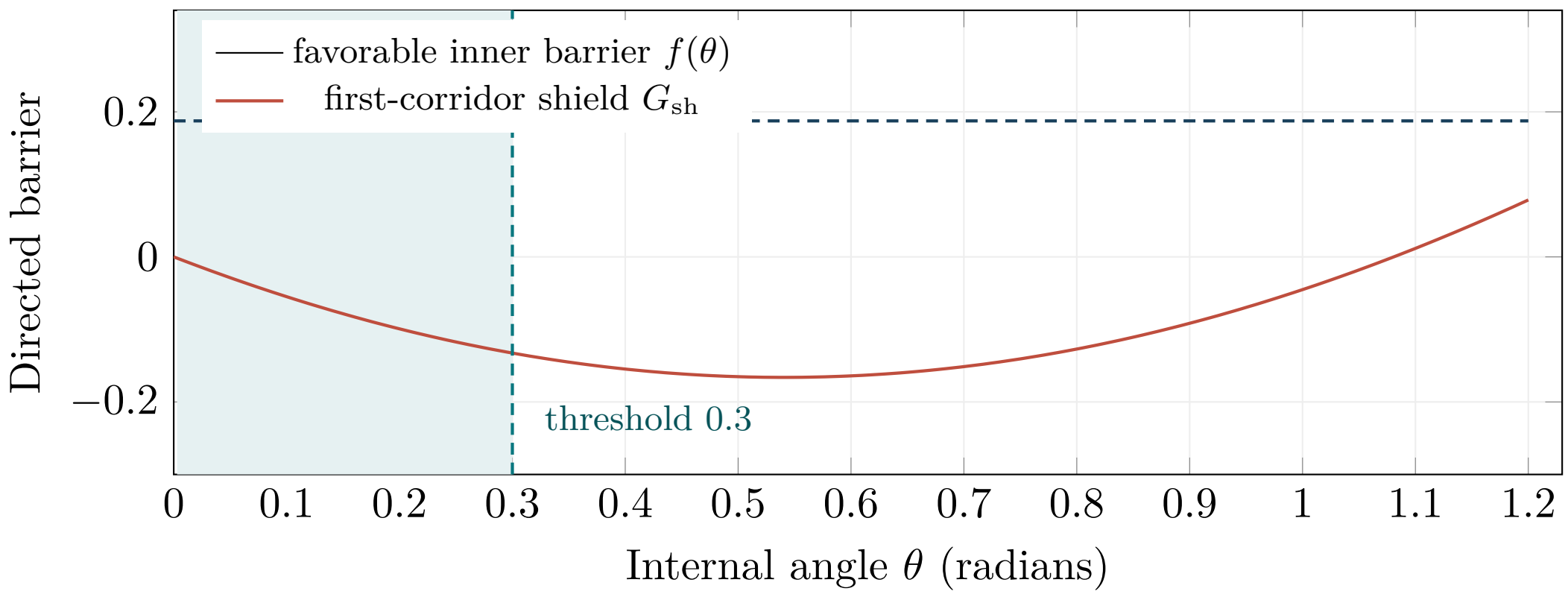

Figure~\ref{fig:collapse-barrier} displays the sign change of the isolated inner barrier and the limited scope of the initial corridor shield.

\subsection{Why this still does not establish the full second prefactor}
The two lemmas make the remaining analytical task narrower, not automatic. For the original three-dimensional flow one must construct a moving angular chart and prove that its \emph{actual} coefficients differ from the isolated $p,q$ in a manner absorbed by $r_\theta,r_\varphi$ with \eqref{eq:regional-ratios}. A bound on $\alpha$ alone does not control the absolute common phase: a simultaneous rotation leaves $\alpha$ unchanged. After collapse, one must also control the \emph{tangential} drift between the full block center and the exact weighted quotient through the second event. The first-corridor shield $G_{\rm sh}$ cannot be used unchanged over that entire trajectory.

In particular, a tentative center-equation remainder should retain potential contributions $O(\beta\|\zeta\|^2)$ from internal nonreciprocal weights, rather than assuming an unverified coefficient-free $O(\|\zeta\|^2)$ bound for the relative displacement $\zeta$. A time-integrated, phase-dependent bound on that remainder and the competing event speeds is the next proof checkpoint. Consequently, the full-flow limit in \eqref{eq:full-target} remains a \emph{conjectural identification}, despite the precise isolated-pair and quotient calculations.

\FloatBarrier
\section{A sharp second exponent without prefactor shadowing}\label{sec:sharp-second}
We establish matching logarithmic upper and lower bounds for two-cluster
interaction times using positive interactions and invariant cones. The
argument does not infer a prefactor from normal proximity.
The cone and projective-coordinate steps below use the positive-interaction
mechanism in~\cite[Lemma~4.2 and its proof]{Perspective} and~\cite{Lifting}.
The sharp clock conclusion requires the separate barrier estimates proved
here; it is not a consequence of consensus alone.

\begin{lemma}[Positive projective coordinates]\label{lem:projective}
Consider $\dot u_i=\sum_jp_{ij}(u_j-(u_i^\top u_j)u_i)$ with continuous
$p_{ij}\ge0$ and $\sum_jp_{ij}=1$. If a unit vector $e$ satisfies
$a_i(0):=e^\top u_i(0)\ge a_*>0$, then $a_i(t)\ge a_*$.
For any fixed linear map $L$, the coordinates $z_i=Lu_i/a_i$ obey
\[
 \dot z_i=\sum_jp_{ij}\frac{a_j}{a_i}(z_j-z_i).
\]
Consequently their convex hull is nonincreasing in time, and
$\max_i\|z_i(t)\|\le\max_i\|z_i(0)\|$.
\end{lemma}
\begin{proof}
At an active minimum of the $a_i$, put $b_i=\sum_jp_{ij}u_i^\top u_j\le1$.
Then $\dot a_i=\sum_jp_{ij}a_j-b_i a_i\ge(1-b_i)a_i\ge0$.
The finite-minimum derivative rule gives invariance of $a_i\ge a_*$.
Differentiating $Lu_i/a_i$ cancels the $b_i$ terms and yields the displayed
equation. Every supporting linear functional has nonincreasing maximum and
nondecreasing minimum for this cooperative equation, proving convex-hull
invariance. This is the standard projective consensus mechanism, related to
the cone-collapse argument of~\cite{Perspective} and the directed-sphere
analysis of~\cite{Lifting}; no priority is claimed for the coordinate lemma.
\end{proof}

\begin{theorem}[Zero-skew two-cluster clock]\label{thm:zero-skew-clock}
Fix a finite number of particles, positive masses, a skew matrix $A$, and
$\varepsilon\in\mathbb R$. Let $v,w\in\mathbb S^{d-1}$ satisfy
\[
 v^\top w=\cos\phi,\qquad 0<\theta_*<\phi<\pi/2,
 \qquad v^\top Aw=0.
\]
For each $\beta$, initialize the smooth normalized interaction flow in two
nonempty blocks $I,J$ such that
\[
 r_\beta:=\max\{\max_{i\in I}\|u_i(0)-v\|,
                       \max_{j\in J}\|u_j(0)-w\|\}\longrightarrow0.
\]
Let $T_\beta$ be the first crossing of angle $\theta_*$ by a pair with one
index in each block. Then
\[
 \boxed{\lim_{\beta\to\infty}\beta^{-1}\log T_\beta
                  =G:=1-\cos\phi.}
\]
No $o(\beta^{-1})$ initial error, transverse eigenvalue assumption, or
full-to-quotient prefactor identification is required.
\end{theorem}
\begin{proof}
Write $K=\|I+\varepsilon A\|_{\rm op}$.
For the lower bound, fix $0<a<G$ and a sufficiently small $\delta>0$.
Initially each block lies in the angular $\delta$-cap about its own center
for all sufficiently large $\beta$. Bootstrap both blocks in their respective
$2\delta$-caps. Since $v^\top Aw=0$, every cross-block barrier in this tube is
at least $G-4K\delta$. The self term in each softmax denominator implies that
the total external weight in any row is at most
$\eta_\beta=C\exp[-\beta(G-4K\delta)]$.
At a minimum of $v^\top u_i$ within $I$, every internal term is nonnegative,
as long as that minimum is positive; external terms have absolute value at
most $2\eta_\beta$. Hence
\[
 \min_{i\in I}v^\top u_i(t)\ge\cos\delta-2\eta_\beta t.
\]
The same estimate holds for the $w$-block. Choose $\delta$ with
$a<G-4K\delta$ and $\phi-4\delta>\theta_*$. On
$0\le t\le e^{\beta a}$, the right-hand error tends to zero, so a first-exit
argument preserves both $2\delta$-caps. Every cross angle exceeds $\theta_*$;
thus $T_\beta\ge e^{\beta a}$. Let $a\uparrow G$.

For the upper bound, let $E=\operatorname{span}\{v,w\}$ and
$e=(v+w)/\|v+w\|$. At the initial time $e^\top u_i\ge a_*>0$ uniformly for
large $\beta$. Apply Lemma~\ref{lem:projective} with $L=P_{E^\perp}$.
The initial normal coordinates are $O(r_\beta)$, hence
$\|P_{E^\perp}u_i(t)\|\le C r_\beta$ at all times. The restriction of $A$ to
$E$ vanishes as a bilinear form: its values on the basis $(v,w)$ are zero by
skew symmetry and $v^\top Aw=0$. Decomposition into $E$ and $E^\perp$ gives
\[
 |u_i(t)^\top Au_j(t)|\le C r_\beta\quad\hbox{for all }i,j,t.
\]
Initially $\rho_\beta(0)=\min_{i,j}u_i(0)^\top u_j(0)=\cos\phi+O(r_\beta)>0$.
The active-pair calculation for acute consensus yields
$\rho_\beta(t)\ge\rho_\beta(0)$. All logits before multiplication by $\beta$
therefore lie between $\rho_\beta(0)-C r_\beta$ and $1+C r_\beta$.
Consequently every transition probability is at least
\[
 k_\beta:=\frac{m_{\min}}{\sum_jm_j}
   \exp\{-\beta[1-\rho_\beta(0)+C' r_\beta]\}.
\]
For an active minimizing pair,
$\dot\rho_\beta\ge2k_\beta(1-\rho_\beta^2)$ almost everywhere.
Integration to $\cos\theta_*$ bounds the time by $C/k_\beta$, where $C$ is
uniform for large $\beta$. In particular all cross pairs have reached the
threshold by that time. Thus
$\limsup\beta^{-1}\log T_\beta\le1-\cos\phi$.
Together with the lower bound this proves the claim.
\end{proof}

\begin{remark}[Exactly what is and is not covered]
The blocks may have any fixed positive combined masses and need not collide
exactly. The zero-skew and acute-angle hypotheses are substantive: the proof
does not cover a general directed second barrier, a right or obtuse limiting
angle, an unbounded number of particles, or a relative prefactor. In particular
the theorem cannot be substituted for the unresolved tangential matching
estimate in a full recursive merger tree.
\end{remark}

\begin{lemma}[One-sided entrance in the explicit three-point example]\label{lem:one-sided}
Use $A=\left(\begin{smallmatrix}0&-1&0\\1&0&0\\0&0&0\end{smallmatrix}\right)$,
$\varepsilon=0.6$, equal masses, and the initial points
\[
 v=(1,0,0),\quad w=(\cos0.8,0,\sin0.8),\quad
 u_3(0)=(\cos1.2,\sin1.2,0).
\]
For each fixed $\delta\in(0,0.3)$, let $s_{\beta,\delta}$ be the first time
$\angle(u_1,u_3)=\delta$. Then
\[
 \beta^{-1}\log s_{\beta,\delta}\longrightarrow B_0,
 \quad u_1(s_{\beta,\delta})\to v,\quad u_2(s_{\beta,\delta})\to w,
 \quad\operatorname{dist}_{\mathbb S}(u_3(s_{\beta,\delta}),v)\le\delta+o(1),
\]
where $B_0=1-\cos1.2-0.6\sin1.2$.
\end{lemma}
\begin{proof}
The same off-diagonal corridor shield applies with threshold $\delta$:
\[
 G_{\rm sh}=1-\cos0.8\sqrt{1+0.6^2}>B_0.
\]
Every competing angular gap is at least $0.8$.
The corridor bootstrap gives the first clock, a tube error
$o(\beta^{-1})$ that is in fact exponentially small for fixed $\delta$, and
an exponentially small time integral of the external row weights.

To establish the additional anchoring assertion, write
$b=1-u_1^\top u_3\ge0$ and $a=-\varepsilon u_1^\top Au_3$.
Inside that tube, while $\theta=\angle(u_1,u_3)\in[\delta,1.2]$,
$a=0.6\sin\theta+o(\beta^{-1})\ge0.3\sin\delta$ for large $\beta$.
After division by the self terms put
$x=e^{-\beta(b+a)}$, $y=e^{-\beta(b-a)}$ and let $s_1,s_3$ denote the
external contributions to the two row denominators. The shield gives
$s_1,s_3=o(1)$, and
\[
 \frac{p_{13}}{p_{31}}=
 \frac{x(1+y+s_3)}{y(1+x+s_1)}
 \le e^{-2\beta a}+e^{-\beta(b+a)}+s_3e^{-2\beta a}
 \le C e^{-\beta a}.
\]
If $q=p_{13}+p_{31}$, then
$\dot\theta=-q\sin\theta+O(\eta_\beta)$ and
$\int_0^{s_{\beta,\delta}}q\sin\theta\,dt\le1.2+C\int\eta_\beta dt=O(1)$.
Therefore the internal contribution to the displacement of $u_1$ is at most
$C e^{-0.3\beta\sin\delta}$; its external contribution tends to zero as well.
The displacement of $u_2$ is entirely external and tends to zero. The triangle
inequality gives the last assertion.
\end{proof}

\begin{corollary}[The full-flow second exponent]\label{cor:full-second}
Let $t_2(\beta)$ be the first time either $(1,2)$ or $(2,3)$ reaches angle
$0.3$ in this example, without forced collisions. Then
\[
 \boxed{\lim_{\beta\to\infty}\frac{\log t_2(\beta)}{\beta}
              =1-\cos0.8\approx0.30329329.}
\]
\end{corollary}
\begin{proof}
The entrance lemma supplies, for each fixed $\delta$, two blocks in
$O(\delta)$-neighborhoods of $v,w$ at a time of logarithmic exponent $B_0$.
Moreover $v^\top Aw=0$ and $\angle(v,w)=0.8<\pi/2$.
For any $0<a<G=1-\cos0.8$, choose $\delta$ sufficiently small and apply the
cap argument of Theorem~\ref{thm:zero-skew-clock} from the entrance time:
the cross barriers are at least $G-C\delta$, and the two blocks remain
separated for a further $e^{\beta a}$ time. This gives $\liminf\beta^{-1}
\log t_2\ge G$. For the upper bound, the projective argument of that theorem,
now with initial normal error $O(\delta)+o(1)$, bounds the remaining time by
$C_\delta e^{\beta(G+C'\delta+o(1))}$. Adding the entrance time contributes
exponent at most $B_0<G$. Let $\delta\downarrow0$ to obtain the matching
upper bound. There is no earlier cross event before entrance because the
corridor keeps the external angles above $0.8-o(1)>0.3$.
\end{proof}

The constant $0.95054656$ for $\beta t_2e^{-\beta G}$ remains an unproved
full-flow prefactor prediction. The corollary resolves the logarithmic
exponent, not that stronger limit. In particular a vanishing angular error
can suffice for this exponent while an $O(\beta^{-1})$ tangential error can
still change the prefactor by an order-one factor.

\section{Growing Systems Governed by Block Mass}\label{sec:block-growth}
Particle count can be separated from the metastable clock more sharply than
a minimum-individual-mass estimate permits. Let $I_\beta,J_\beta$ be two
nonempty blocks, with total masses $M_I,M_J$, and define
$R_\beta=\max(M_I/M_J,M_J/M_I)$. The number of particles in each block is
arbitrary and finite for each $\beta$.

\begin{theorem}[Block-mass clock]\label{thm:block-mass}
Fix $0<\theta_*<\pi/2$ and positive individual masses.
Let $\varepsilon$ be fixed, $A_\beta^\top=-A_\beta$,
$\sup_\beta\|A_\beta\|_{\rm op}<\infty$, and
let unit centers $v_\beta,w_\beta$ satisfy
$v_\beta^\top A_\beta w_\beta=0$ and
$\angle(v_\beta,w_\beta)\to\phi\in(\theta_*,\pi/2)$.
Suppose the maximal initial distance from each particle to its block center
is $r_\beta\to0$, and
\[
 \beta^{-1}\log R_\beta\longrightarrow0.
\]
Then the first cross-block angle-$\theta_*$ event satisfies
\[
 \boxed{\lim_{\beta\to\infty}\beta^{-1}\log T_\beta=1-\cos\phi.}
\]
No bound on $\beta^{-1}\log |I_\beta\cup J_\beta|$ or on individual mass
ratios is required. In particular, balanced blocks may contain
$\lceil e^{\gamma\beta}\rceil$ particles, for any fixed $\gamma>0$.
\end{theorem}

\begin{proof}
All constants below depend only on a compact neighborhood of the limiting
angle, the fixed threshold, and the uniform operator-norm bound; they do not
depend on particle count or dimension. Put $G=1-\cos\phi$.

\emph{Lower bound using total block mass.}
Fix $a<G$ and a small fixed $\delta>0$. While each block stays in the angular
$2\delta$-cap about its center, a row's internal scores differ from 1 by at
most $C\delta$, and cross scores differ from $\cos\phi$ by at most
$C\delta+o(1)$. Summing the entire internal block in the denominator gives
\[
 \sum_{j\notin I_\beta}p_{ij}
 \le R_\beta e^{-\beta(G-C\delta-o(1))},\qquad i\in I_\beta,
\]
and the analogous estimate holds for $J_\beta$. At the minimum of the
center scalar product within a block, internal terms are nonnegative and
the external contribution is bounded below by minus twice this row weight.
Choose $\delta$ with $a<G-C\delta$ and $\phi-4\delta>\theta_*$.
The integrated external force to time $e^{\beta a}$ tends to zero, since
$\log R_\beta=o(\beta)$. The cap first-exit argument then closes, giving
$T_\beta\ge e^{\beta a}$.

\emph{Subexponential entrance to small block diameter.}
Set $\alpha_\beta=\sqrt{r_\beta}+\beta^{-1/4}$. Bootstrap the two blocks
inside $2\alpha_\beta$-caps. Their external row weights satisfy
\[
 \eta_\beta\le R_\beta e^{-\beta(G-O(\alpha_\beta)-o(1))}
                         =e^{-\beta G+o(\beta)}.
\]
For the $v$-block use $a_i=v^\top u_i$ and $z_i=u_i/a_i$.
Inside these caps, $a_i\ge1/2$. Its internal projective coefficients obey
\[
 p_{ij}\frac{a_j}{a_i}\ge\kappa_\beta\frac{m_j}{M_I},
 \qquad\kappa_\beta=c e^{-C\beta\alpha_\beta},\qquad i,j\in I_\beta.
\]
Indeed, all internal score differences are $O(\alpha_\beta)$ and the external
numerator is negligible relative to the full internal denominator. The
external projective forcing has norm at most $C\eta_\beta$; the other
block's projective coordinates are bounded because $\phi<\pi/2$.

Let $D_I=\max_{i,j\in I}\|z_i-z_j\|$. At an active nonzero diameter,
its direction is a supporting direction at both endpoints. Subtract the
common positive kernel $\kappa_\beta m_j/M_I$ from each row. Remaining
internal coefficients are nonnegative and cannot enlarge that supporting
range. This is a common-kernel consensus contraction, consistent with the
positive-system approach of~\cite{Moreau}. The common kernel contracts the range at rate $\kappa_\beta$.
Thus, in upper Dini derivatives,
\[
 D_I^+\le-\kappa_\beta D_I+C\eta_\beta,
 \quad D_I(t)\le D_I(0)e^{-\kappa_\beta t}+C\eta_\beta/\kappa_\beta.
\]
The same estimate holds for $J$. The inverse map $z\mapsto z/\|z\|$
is uniformly Lipschitz on $v^\top z=1$, so these projective diameter bounds
also control spherical-particle diameters. Take
$s_\beta=\beta/\kappa_\beta=e^{o(\beta)}$.
Then $D_I(s_\beta)+D_J(s_\beta)\le e^{-c_0\beta}$ for a fixed $c_0>0$.
The cap bootstrap is valid to that time: the integrated external force is
$e^{-\beta G+o(\beta)}$, much smaller than the cap boundary margin
$\cos(r_\beta+o(r_\beta))-\cos(2\alpha_\beta)$, which is bounded below by
a fixed multiple of $\alpha_\beta^2\ge\beta^{-1/2}$ for large $\beta$.

\emph{Persistence until the cross event.}
The global bisector projective argument bounds all components normal to
$\operatorname{span}(v,w)$ by $C r_\beta$ for all times. The restriction of
$A_\beta$ to this plane vanishes, so all skew pairings are $O(r_\beta)$.
Before the first cross event, cross scalar products are at most
$\cos\theta_*$. With the blocks already exponentially narrow, every query
on a segment joining two same-block particles consequently has external
weight at most
$R_\beta e^{-\beta(1-\cos\theta_*-o(1))}$.

For $V_{ij}=1-u_i^\top u_j$ within a block let
$V=\max_{i,j}V_{ij}$. Freeze the configuration and write
$\mu(x)=\sum_\ell p_\ell(x)u_\ell$ and $\Delta=u_i-u_j$.
Direct differentiation gives the two identities
\[
 \dot V_{ij}=\Delta^\top[\mu(u_i)-\mu(u_j)]
       -V_{ij}[u_i^\top\mu(u_i)+u_j^\top\mu(u_j)],
 \qquad D_x\mu=\beta\operatorname{Cov}_{p(x)}(u)M^\top.
\]
On each segment the covariance trace is bounded by the second moment
about any same-block key, giving
$\|\operatorname{Cov}_{p(x)}(u)\|_{\rm op}\le2V+4\eta$.
At either endpoint, internal keys have scalar product at least $1-V$,
so $u_i^\top\mu(u_i)\ge1-V-2\eta$. The mean-value formula and
$\|\Delta\|^2=2V_{ij}$ therefore give, at an active maximum,
\[
 D^+V\le[-2+2V+4\eta+4\beta K V+8\beta K\eta]V,
 \qquad K=\|I+\varepsilon A_\beta\|_{\rm op}.
\]
At $s_\beta$ both $\beta V$ and $\beta\eta$ tend to zero. A first-exit
argument makes the bracket at most $-1$ up to the cross event, so the block
diameters remain exponentially small throughout this interval.

\emph{Matching upper clock.}
The initial global Gram minimum is $\rho(0)=\cos\phi+o(1)>0$ and remains
nondecreasing. Before the first cross event an active global minimum is a
cross pair, since internal diameters are exponentially small. For an active
pair $i\in I,j\in J$, retain every $\ell\in J$ in its first Gram sum and
every $\ell\in I$ in its second. Internal narrowness yields
\[
 \dot\rho\ge(P_{iJ}+P_{jI})(1-\rho^2-o(1)).
\]
The remaining terms are nonnegative by the active-minimum argument. All
scores lie between $\cos\phi-o(1)$ and $1+o(1)$; hence
\[
 P_{iJ}\ge\frac{M_J}{M_I+M_J}e^{-\beta(G+o(1))},\qquad
 P_{jI}\ge\frac{M_I}{M_I+M_J}e^{-\beta(G+o(1))}.
\]
The mass fractions add to one. While $\rho\le\cos\theta_*$, the last factor
in the derivative estimate is at least $\sin^2\theta_*/2$ for large $\beta$.
Integration bounds the remaining time by $C e^{\beta(G+o(1))}$. Adding the
subexponential entrance time $s_\beta$ proves the upper bound and completes
the proof.
\end{proof}

\begin{corollary}[Count invariance and an exponential-imbalance boundary]
Exact replication of any particle, with its mass split among the copies,
leaves every distinct trajectory unchanged. More generally, arbitrarily
growing balanced near-collision blocks satisfy Theorem~\ref{thm:block-mass}.
For exactly collapsed zero-skew blocks with
$M_J/M_I=e^{\beta a}$, $a\in\mathbb R$, the exponent instead equals
$[1-\cos\phi-|a|]_+$.
\end{corollary}
\begin{proof}
Equal queries have equal velocities, and their numerator weights add to the
original mass, proving exact replication invariance by uniqueness. For two
exact blocks, the quotient probabilities are
\[
 q_{IJ}=\frac1{1+e^{\beta(1-\cos\psi-a)}},\qquad
 q_{JI}=\frac1{1+e^{\beta(1-\cos\psi+a)}}.
\]
The angular hitting integral has inverse-rate exponent
$[1-\cos\psi-|a|]_+$, whose maximum on $[\theta_*,\phi]$ is its value at
$\phi$. The endpoint Laplace principle gives the displayed rate. This also
shows why a bound based on the smallest individual mass confounds particle
replication with physical block imbalance.
\end{proof}

\section{Uniformity, Growing Systems, and Verification Details}
\begin{theorem}[Dimension-free triangular-array clock]\label{thm:growing-clock}
Fix $0<\theta_*<\pi/2$.
The zero-skew two-cluster clock theorem remains valid for a family with
$N=N_\beta$ particles, dimension $d=d_\beta$, and positive masses
$m_{i,\beta}$ and skew matrices $A_\beta$, provided
\[
 \frac1\beta\log\frac{\sum_i m_{i,\beta}}{\min_i m_{i,\beta}}\to0,
 \quad\sup_\beta\|A_\beta\|_{\rm op}<\infty,
 \quad v_\beta^\top A_\beta w_\beta=0,
 \quad\angle(v_\beta,w_\beta)\to\phi\in(\theta_*,\pi/2),
\]
and the maximal initial block radius tends to zero. Here $\varepsilon$ is
fixed. Its conclusion is $\beta^{-1}\log T_\beta\to1-\cos\phi$.
\end{theorem}
\begin{proof}
Fix $a<1-\cos\phi$ and choose a fixed sufficiently small cap radius $\delta$.
The cross-barrier estimate is uniform because the operator norms are bounded.
The total external row weight is at most
\[
 \frac{\sum_i m_{i,\beta}}{\min_i m_{i,\beta}}
       e^{-\beta(1-\cos\phi-O(\delta)-o(1))}.
\]
Its integral to $e^{\beta a}$ tends to zero by the subexponential mass-ratio
assumption. The cap first-exit argument is unchanged and gives the lower bound.

For the upper bound, choose the bisector of the two centers. Its scalar
products with all initial particles exceed a common constant $a_*>0$, since
$\phi<\pi/2$ and the initial block radii tend to zero. The projective
normal-coordinate argument bounds every normal component by $r_\beta/a_*$;
convex-hull invariance and operator-norm estimates have no factors of $N$ or
$d$. Thus all skew pairings are $O(r_\beta)$ uniformly. The initial minimum
Gram entry is $\cos\phi+o(1)$ and stays nondecreasing. The row-wise transition
lower bound is
\[
 k_\beta=\frac{\min_i m_{i,\beta}}{\sum_i m_{i,\beta}}
          e^{-\beta(1-\cos\phi+o(1))}.
\]
The acute comparison integrates explicitly as
\[
 T_\beta\le
 \frac{\operatorname{arctanh}(\cos\theta_*)-
             \operatorname{arctanh}(\rho_\beta(0))}{2k_\beta}
\]
when $\rho_\beta(0)<\cos\theta_*$; the numerator is bounded uniformly for
large $\beta$. Taking logarithms and using the mass-ratio assumption gives
the matching upper rate. For each fixed $\beta$, only a finite minimum is
taken, so the standard absolutely continuous minimum rule applies even
though the number of coordinates varies with $\beta$.
\end{proof}

The mass condition cannot simply be discarded. For two zero-skew centers
with initial angle $\phi$, let $m_2/m_1=e^{\beta a}$, where
$0<a<1-\cos\phi$. The two-body angular integral has rate
$[1-\cos\phi-a]_+$, because the favorable transition contains the exponent
$1-\cos\psi-a$. Its maximal positive value on the crossing interval occurs
at $\phi$. Thus exponential mass imbalance changes the clock even though
the centers and skew pairing are unchanged.

For completeness, the acute Gram minimum used above satisfies the following
direct inequality. At a minimizing off-diagonal pair $(i,j)$ with $c_{ij}=\rho\ge0$,
\[
 \dot c_{ij}=\sum_\ell p_{i\ell}(c_{\ell j}-c_{i\ell}\rho)
            +\sum_\ell p_{j\ell}(c_{i\ell}-c_{j\ell}\rho).
\]
Each summand is nonnegative since $c_{\ell j}\ge\rho$ and $c_{i\ell}\le1$.
Keeping $\ell=j$ and $\ell=i$ gives
$\dot\rho\ge(p_{ij}+p_{ji})(1-\rho^2)\ge2k_\beta(1-\rho^2)$ almost
everywhere. This proves both invariance and the integrated bound, with no
gradient-flow assumption.

The numerical verification compares the projected normal equation against
an explicitly assembled skew-basis linear system, compares its derivative
with automatic differentiation and centered finite differences, and tests
the second-order attention bound over multiple dimensions, temperatures,
skew strengths, masses, and radii. These tests check the implemented
equations and constants. The analytic theorems are not claimed to be fully
formalized by the finite-dimensional algebra lemmas.

\section{A reproducible integration specification}
For validation, set $Y(t)=(u_1(t),u_2(t),u_3(t))\in\R^9$ and evaluate each row softmax as
\begin{equation}
s_{ij}=\beta u_i^\T M u_j,\quad a_i=\max_j s_{ij},\quad
p_{ij}=\frac{e^{s_{ij}-a_i}}{\sum_k e^{s_{ik}-a_i}}.
\end{equation}
Integrate $\dot u_i=\sum_jp_{ij}[u_j-(u_i^\T u_j)u_i]$ with DOP853 using relative tolerance $10^{-9}$ and absolute tolerance $10^{-11}$. Trigger a crossing when $u_i^\T u_j-\cos(0.3)$ changes from negative to positive, adjusting for floating-point drift by normalizing the dot product in the event function. Continue without modifying the state after the first event; the next previously untriggered pair defines the second event. The numerical observations concern only the specified finite parameter values.

\begingroup\small\raggedright

\endgroup
\end{document}